\documentclass[letterpaper]{article} 
\usepackage[utf8]{inputenc}
\usepackage[]{aaai2026}  
\usepackage{times}  
\usepackage{helvet}  
\usepackage{courier}  
\usepackage[hyphens]{url}  
\usepackage{graphicx} 
\usepackage{natbib}  
\usepackage{caption} 
\usepackage{algorithm}
\usepackage{algorithmic}
\usepackage{color}
\usepackage{colortbl}
\newcommand{\dec}[1]{{\small\hspace{0.05cm}{\color[HTML]{CD5C5C}{$_{\textbf{-#1}}$}}}}

\newcommand{\imp}[1]{{\small\hspace{0.05cm}{\color[HTML]{32CB00}{$_{\textbf{+#1}}$}}}}

\usepackage{tabularx, array}
\newcolumntype{C}[1]{>{\centering\arraybackslash}p{#1}}
\usepackage{hyphenat}
\newcolumntype{Y}{>{\centering\arraybackslash}X} 
\usepackage[most]{tcolorbox}
\tcbuselibrary{breakable} 
\usepackage{fancyvrb}    
\tcbuselibrary{breakable,skins}   
\tcbuselibrary{listings, breakable}
\usepackage{enumitem}  
\usepackage{fvextra}   
\usepackage{booktabs}
\usepackage{pifont}
\usepackage{multirow}
\DefineVerbatimEnvironment{Verbatim}{Verbatim}{breaklines=true, breakanywhere=true}

\definecolor{ysdarkpurple}{HTML}{4E2399}
\definecolor{ysshallowpurple}{HTML}{E6DBFF}
\definecolor{ysdarkred}{HTML}{8c2824}
\definecolor{ysshallowred}{HTML}{F8D7D7}
\definecolor{ysdarkblue}{HTML}{005E99}
\definecolor{ysshallowblue}{HTML}{CCEBFF}
\definecolor{ysdarkgrey}{HTML}{333333}
\definecolor{ysshallowgrey}{HTML}{E5E5E5}

\newtcbinputlisting{\promptbox}[2][]{
  enhanced,
  breakable, 
  colback=ysshallowblue,
  colframe=ysdarkblue,
  fonttitle=\bfseries,
  title=#2,
  listing only, 
  listing options={
    language=YAML,          
    basicstyle=\ttfamily\small, 
    breaklines=true,        
    breakatwhitespace=true, 
    postbreak=\mbox{\textcolor{red}{$\hookrightarrow$}\space}, 
    showstringspaces=false, 
    upquote=true,           
  },
  #1 
}

\definecolor{ColorGrok}{HTML}{FFFDE7}      
\definecolor{ColorPplx}{HTML}{EFFDFE}      
\definecolor{ColorOpenAI}{HTML}{F2F2F2}    
\definecolor{ColorGemini}{HTML}{E6F4FE}    
\definecolor{ColorClaude}{HTML}{FFF3EB}    
\definecolor{SectionHeaderColor}{HTML}{FFFFFF} 

\colorlet{DarkerColorClaude}{ColorClaude!95!black}
\colorlet{DarkerColorPplx}{ColorPplx!95!black}
\colorlet{DarkerColorGemini}{ColorGemini!95!black}
\colorlet{DarkerColorOpenAI}{ColorOpenAI!95!black}
\colorlet{DarkerColorGrok}{ColorGrok!95!black}
\graphicspath{{figures/}}

\usepackage{newfloat}
\usepackage{listings}
\DeclareCaptionStyle{ruled}{labelfont=normalfont,labelsep=colon,strut=off} 
\floatstyle{ruled}
\newfloat{listing}{tb}{lst}{}
\floatname{listing}{Listing}
\title{\ \ \ \ Beyond Brainstorming: What Drives High-Quality Scientific Ideas? \\ Lessons from Multi-Agent Collaboration}
\author{
\ \ Nuo Chen \quad
\ Yicheng Tong \quad
Jiaying Wu \ \ 
Minh Duc Duong\\
Qian Wang \quad
Qingyun Zou \quad
Bryan Hooi \quad
Bingsheng He  \vspace{2mm}\\
}
\affiliations{
    \large  \textsuperscript{\rm }National University of Singapore\\

}

\usepackage{bibentry}

\begin{document}

\maketitle

\footnotetext[1]{
\hspace{\footnotesep} nuochen@comp.nus.edu.sg \\ \hspace*{2.5em} Proposal Evaluation Demo: \url{https://rateyourproposal.ai} \\ \hspace*{2.5em} Code: \url{https://github.com/NuoJohnChen/Idea2Proposal}}

\begin{abstract}
While AI agents show potential in scientific ideation, most existing frameworks rely on single-agent refinement, limiting creativity due to bounded knowledge and perspective. Inspired by real-world research dynamics, this paper investigates whether structured multi-agent discussions can surpass solitary ideation. We propose a cooperative multi-agent framework for generating research proposals and systematically compare configurations including group size, leader-led versus leaderless structures, and team compositions varying in interdisciplinarity and seniority. To assess idea quality, we employ a comprehensive protocol with agent-based scoring and human review across dimensions such as novelty, strategic vision, and integration depth. Our results show that multi-agent discussions substantially outperform solitary baselines. A designated leader acts as a catalyst, transforming discussion into more integrated and visionary proposals. Notably, we find that cognitive diversity is a primary driver of quality, yet expertise is a non-negotiable prerequisite, as teams lacking a foundation of senior knowledge fail to surpass even a single competent agent. These findings offer actionable insights for designing collaborative AI ideation systems and shed light on how team structure influences creative outcomes.

\end{abstract}

\section{Introduction}

Idea generation is a foundational driver of scientific discovery, transforming initial curiosity into meaningful breakthroughs. While some advances originate from individual insight, history shows that many of the most impactful innovations emerge through collaboration, where diverse perspectives converge through discussion, debate, and synthesis~\cite{paulus2003group, nijstad2006group}. Despite this, most AI-driven ideation frameworks remain centered on single-agent systems that refine ideas in isolation, lacking the dynamic exchange of multiple viewpoints. Although efficient, this solitary approach carries intrinsic limitations from the underlying models, including incomplete knowledge, susceptibility to confirmation bias, and restricted creativity. Such creativity often arises from the interplay of contrasting opinions and the integration of complementary insights~\cite{sweller1988cognitive, page2007difference}.

\begin{figure}
    \centering
    \includegraphics[width=\linewidth]{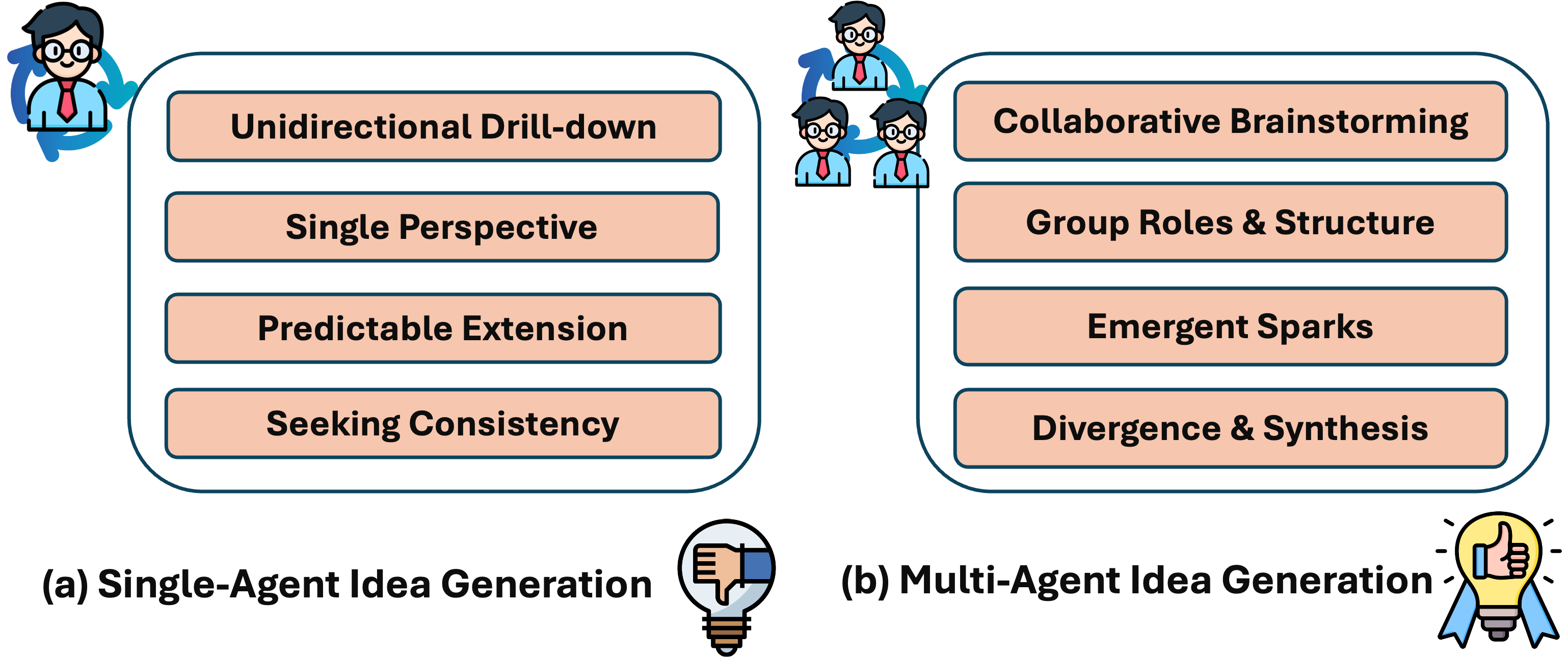}
    \caption{Comparison of single-agent and multi-agent scientific ideation across interaction modes, agent composition, sources of innovation, and core mechanisms.}

    \label{fig:single_multi}
\end{figure}

\begin{table*}[h]
\centering
\caption{Comparison with prior approaches. Our method uniquely combines agent heterogeneity, structured collaboration, and collaboration-sensitive evaluation (CSE). \ding{52} indicates the presence of CSE.}

\label{tab:approach_comparison}
\resizebox{\textwidth}{!}{%
\begin{tabular}{@{}lclllc@{}}
\toprule
\textbf{Paper / Approach} & \textbf{\begin{tabular}[c]{@{}l@{}}\# of \\ Agents\end{tabular}} & \textbf{Agent Diversity} & \textbf{\begin{tabular}[c]{@{}c@{}}Mode of\\ Communication\end{tabular}}  &  \textbf{Output}   & \textbf{CSE} \\ \midrule
AI-Researcher \cite{Si2025Can}                                                  & 1                     & Prompt-guided
self-reflection                      & Internal refinement loops                                                                   & Idea Snippet & 
\ding{55} \\
   The AI Scientist \cite{lu2024aiscientistv1}                & 1                     & Internal roles + planner     & Sequential task handoff                                             & Full Paper & \ding{55} \\
The AI Scientist-v2 \cite{yamada2025aiscientistv2}                 & 1                     & Internal roles
+ tree search     & Hierarchical task decomposition                                             & Full Paper & 
\ding{55}  \\
CycleResearcher \cite{weng2025cycleresearcherimprovingautomatedresearch}          & 1                     & Role separation via reward design   & Internal review loop                                                    & Full Paper & 
\ding{55}  \\
ResearchAgent \cite{baek2025researchagentiterativeresearchidea}                 & $1$                    & Knowledge-augmented
single agent    & Sequential refinement                                  & Idea Snippet & 
\ding{55}  \\
VIRSCI  \cite{su2025headsbetteroneimproved}                                 & $>1$                    & Homogeneous    & Simple broadcast                                   & Idea Snippet & \ding{55}  \\
ResearchTown  \cite{yu2024researchtown}                                    &  $>1$                   & Role-based               & Graph-based
message passing                                                    & Full Paper & 
\ding{55}  \\
\midrule
\textbf{Ours} & 
$>1$ & 
\begin{tabular}[l]{@{}l@{}}\textbf{Rich} \\ (Discipline / Seniority)\end{tabular} & 
\begin{tabular}[l]{@{}l@{}}\textbf{Structured} (i.e. Brainstorm +  \\ Lead+ Synthesize)\end{tabular} & 
\begin{tabular}[c]{@{}c@{}}\textbf{Actionable}\\ \textbf{Proposal}\end{tabular} & 
\ding{52}  \\
\bottomrule
\end{tabular}%
}
\end{table*}

Prior work on AI-assisted ideation has primarily focused on single-agent refinement. Systems such as \cite{Si2025Can} and AI Scientist~\cite{weng2025cycleresearcherimprovingautomatedresearch} demonstrate strong capabilities in iterative self-improvement but operate without the interpersonal dynamics that characterize human collaboration. Recent multi-agent approaches~\cite{baek2025researchagentiterativeresearchidea, su2025headsbetteroneimproved} begin to introduce agent interactions, yet often employ limited diversity in agent roles and rely on simplified communication protocols. Table~\ref{tab:approach_comparison} highlights these distinctions, showing how our framework advances the field by incorporating heterogeneous agent compositions—including variations in disciplinary expertise and simulated seniority—and enabling more sophisticated interaction structures such as brainstorming with or without designated leadership. Figure~\ref{fig:single_multi} further contrasts single-agent and multi-agent setups along key dimensions including interaction style, agent composition, sources of innovation, and coordination mechanisms.

This gap elicits a deeper question that extends beyond the mechanics of brainstorming: \textbf{What principles govern the generation of high-quality scientific ideas?}

In this paper, we address the limitations of solitary ideation by developing a multi-agent framework for scientific idea generation, focused on producing high-quality research proposals in AI domains aligned with ICLR topics. Our goal is not only to demonstrate the advantage of multi-agent collaboration but also to systematically investigate the underlying mechanisms that contribute to the generation of high-quality ideas. We draw inspiration from human group dynamics to inform optimal agent cooperation. To this end, we define quality as a composite of five dimensions: novelty, feasibility, impact, coherence, and ethical soundness. These dimensions are assessed using a structured evaluation rubric supported by both automated agent simulations and human expert reviews. Figure~\ref{fig:discussion} illustrates our system, showcasing an example dialogue among agents and the resulting synthesized proposal.

Grounded in theories from organizational psychology and cognitive science~\cite{paulus2003group, MUMFORD2002leading, page2007difference}, we examine how factors such as group size, leadership structure, and team diversity influence ideation outcomes. These principles are instantiated in our framework through agent configurations that simulate variation in research seniority and disciplinary expertise. Our investigation is organized around the following three research questions:

\begin{itemize}
    \item \textbf{RQ1: Group Size and Interaction Scale.} How do pairwise discussions compare to solitary ideation, and how does performance scale in larger group settings?
    \item \textbf{RQ2: Leadership Structure.} What is the effect of having a coordinating leader versus an egalitarian structure on idea quality, particularly in balancing creativity and coherence?
    \item \textbf{RQ3: Agent Composition and Diversity.} How do differences in team composition (e.g.,  interdisciplinary versus intradisciplinary expertise, or mixed versus uniform seniority) affect the quality of generated proposals?
\end{itemize}

Through these questions, we aim to uncover core principles of AI-assisted scientific collaboration. Our contributions are threefold. First, we present a controllable framework for simulating scientific collaboration, showing clear benefits over single-agent baselines. Second, we identify which collaborative configurations most effectively support ideation, connecting insights from social science with AI system design. Third, we propose and validate a comprehensive rubric for evaluating AI-generated scientific proposals. Together, these contributions offer practical guidance for building more creative AI systems and deepen our understanding of how machines and humans can collaborate to generate meaningful scientific ideas.

\section{Theoretical Motivation: Key Cognitive and Social Factors in Group Ideation}
Before developing our framework, it is essential to ground our inquiry in the cognitive and social foundations that shape collaborative creativity. Our decision to pursue multi-agent systems is motivated by established, and at times competing, theories from psychology and organizational science. This section addresses two key questions: \textbf{What cognitive and social mechanisms drive successful group ideation?} And \textbf{How can these insights inform the design of more effective collaborative systems}, whether human, artificial, or hybrid?

\subsection{Cognitive Stimulation and Process Constraints}
A central principle in group creativity is cognitive stimulation: exposure to others’ ideas can activate novel associative pathways, enabling insights that may not emerge in isolation~\cite{paulus2003group, paulus2000groups, nijstad2006group}. This is the primary rationale behind brainstorming. However, collaboration introduces process losses. Production blocking, where participants must wait their turn to contribute, disrupts thought processes and leads to idea loss~\cite{paulus2003group}. Evaluation apprehension, or fear of negative judgment, can inhibit the sharing of unconventional ideas~\cite{nijstad2006group}. These effects motivate a core research question: \textbf{Can a collaborative system overcome its inherent process losses to provide net benefits over solitary ideation?} Optimizing idea generation thus requires balancing stimulation with structure, motivating our exploration of ideal group configurations and interaction protocols~\cite{Coskun2000cognitive}.

\subsection{Leadership Dynamics, Conflict, and Emergence}
Leadership plays a essential yet complex role in creative collaboration. Transformational leaders—those who build trust, encourage autonomy, and inspire collective vision—tend to foster more creative outputs than directive leaders~\cite{MUMFORD2002leading}. However, strong leadership can also suppress dissent unless efforts are made to promote constructive conflict and knowledge integration~\cite{wang2015enhancing}. Evidence suggests that group creativity often \textit{emerges} from negotiating disagreement and integrating opposing viewpoints, rather than merely combining individual contributions~\cite{nijstad2006group}. This motivates an essential investigation: \textbf{In what ways does leadership affect the generation and integration of creative ideas within teams?}

\subsection{Diversity-Induced Creativity and the Coordination Tradeoff}

Extensive research shows that team diversity, whether in expertise, background, or perspective, can broaden idea spaces and increase the likelihood of breakthrough innovation~\cite{paulus2007toward, sauder2016qualitative}. However, diverse teams also face increased challenges in communication, coordination, and trust-building~\cite{page2007difference}. This tension, often referred to as the diversity-innovation tradeoff, highlights a fundamental dilemma: \textbf{How can one balance the benefits of diverse agent composition with the coordination costs that such diversity introduces?} Effective systems must be designed to navigate this tradeoff, rather than simply maximizing one dimension at the expense of the other~\cite{denscombe2012research}.

\subsection{Cognitive Load and Scalability of Interaction}

Collaboration is constrained by cognitive load. Working memory is limited~\cite{sweller1988cognitive}, and excessive interaction complexity can hinder deep processing of others' ideas~\cite{nijstad2002cognitive, nijstad2006group}. Large or unstructured group discussions risk overwhelming participants, leading to shallow engagement and reduced creativity~\cite{paulus2000groups}. These insights motivate our focus on the structure and scale of interactions, particularly the hypothesis that \textbf{moderate group sizes and bounded discussion length} strike a balance between idea diversity and cognitive feasibility.

\subsection{Implications for Multi-Agent Simulation of Ideation}

Together, these theories do not prescribe a single ideal structure for collaboration. Instead, they define a multi-dimensional design space. Our work operationalizes these theoretical insights in a controllable multi-agent framework that simulates scientific ideation. By systematically varying parameters such as group size, communication structure, and agent composition, we aim to uncover design principles that support high-quality idea generation. This framework enables us to map the landscape of collaborative creativity and inform the development of more effective ideation systems: whether human-led, AI-driven, or jointly orchestrated.

\section{Methodology}
\begin{figure*}[h]
    \centering
    \includegraphics[width=0.99\linewidth]{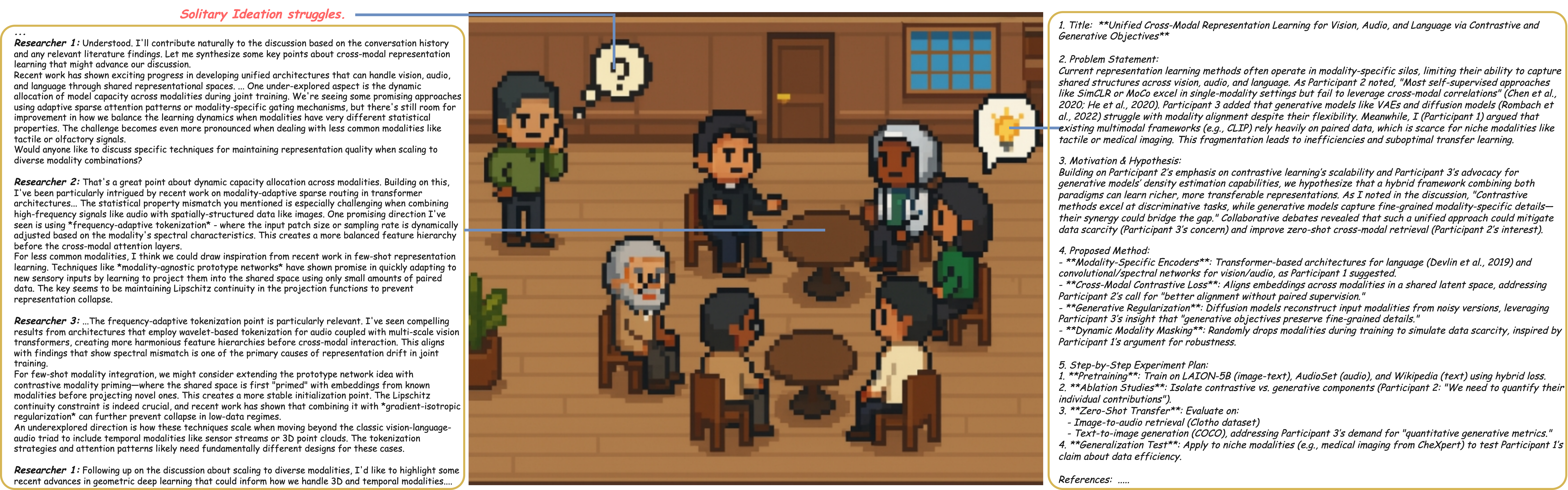}
    \caption{Example of our multi-agent framework in action. \textbf{Left}: Collaborative dialogue among agents. \textbf{Right}: Final synthesized research proposal produced from the discussion.}

    \label{fig:discussion}
\end{figure*}

Leveraging recent progress in large language model (LLM)–based human simulation~\cite{anthis2025llmsocialsimulationspromising, wang2025limits}, we design a multi-agent simulation framework to explore how collaborative structures shape scientific ideation. Our methodology consists of two stages (see Figure \ref{fig:discussion}): (1) a multi-agent discussion phase that generates a research proposal, and (2) a comprehensive evaluation phase that assesses proposal quality. Full prompt templates are included in the Appendix.

\subsection{Idea Generation}

The idea generation pipeline is structured to emulate academic collaboration, guiding agents from open-ended dialogue to a structured proposal.

\paragraph{Collaborative Discussion Phase.}
Given a research topic $T$, a group of agents $A = \{a_1, a_2, \ldots, a_n\}$ engages in a discussion over $R{-}1$ rounds to generate a proposal $P$. The agent configuration (e.g., single agent, leader-led group, interdisciplinary team) varies by experimental condition. In each round $r \in \{1, \ldots, R-1\}$, each agent $a_i$ contributes an utterance $u_{i,r}$ conditioned on (i) topic $T$, (ii) discussion history $H_{r-1} = \{u_{j,k} | j \in \{1..n\}, k < r\}$, and (iii) relevant work.

To ensure the scientific grounding and timeliness of the discussion, agents are equipped with a literature search tool, allowing them to perform queries via a function, $\text{Search}(\text{query})$, which interfaces with the Semantic Scholar API. This enables them to find, verify, and cite relevant, real-world literature to support their arguments.

\paragraph{Proposal Synthesis Phase.}
We use structured research proposals as the final output of ideation. This format captures and integrates key ideas discussed by agents and provides a tractable, evaluable artifact for downstream analysis~\cite{denscombe2012research}. Compared to full-length scientific papers with simulated results~\cite{weng2025cycleresearcherimprovingautomatedresearch}, proposals focus on research framing and planning, avoiding unverifiable claims.

In the final round $R$, a designated agent synthesizes the full discussion history $H_{R-1}$ into a unified research proposal $P$, incorporating citations retrieved during the dialogue. Unlike approaches that generate complete papers with speculative results~\cite{weng2025cycleresearcherimprovingautomatedresearch}, which may compromise scientific validity, our framework produces structured proposals that focus on idea formulation and planning. To ensure comparability across settings, each proposal follows a standardized format grounded in established criteria for high-quality research design~\cite{bordage2001criteria, steven2015criteria}:
\begin{enumerate}
    \item \textbf{Title:} A concise and descriptive title.
    \item \textbf{Problem Statement:} A clear articulation of the current research gap.
    \item \textbf{Motivation \& Hypothesis:} The rationale for the research and the central claim.
    \item \textbf{Proposed Method:} A detailed technical description of the new proposed approach.
    \item \textbf{Experiment Plan:} A concrete plan for validation, including baselines and metrics.
\end{enumerate}

\subsection{Proposal Evaluation}
To evaluate the quality of a generated proposal $P$, we design a holistic framework that captures the influence of collaboration on idea development. The evaluation combines multi-reviewer scoring, reflective refinement, and ensemble aggregation.

\paragraph{Multi-faceted Evaluation Criteria.} 
Our evaluation rubric extends beyond novelty to capture dimensions shaped by collaborative dynamics. Criteria are grouped into two categories and scored on a 1–10 scale per dimension, with an aggregated \textbf{Overall} quality score. Full rubric details are included in the Appendix.

\begin{itemize}
    \item \textbf{Core Quality Dimensions} assess the foundational soundness of a proposal, drawing from established creativity frameworks~\cite{djte:06}, including:  
    (1) \textbf{Novelty}, which captures originality and paradigm-shifting potential, reflecting how multi-agent synthesis can yield unexpected insights;  
    (2) \textbf{Workability}, which evaluates feasibility and sensitivity to real-world constraints, highlighting how collaboration surfaces implementation challenges;  
    (3) \textbf{Relevance}, which measures alignment with the research problem, showing how discussion sharpens topical focus; and  
    (4) \textbf{Specificity}, which assesses clarity and completeness, indicating how iterative exchange improves logical structure.   
    \item \textbf{Collaboration-Sensitive Dimensions} capture qualities uniquely influenced by group interaction, including:  
    (1) \textbf{Integration Depth (ID)}, which measures the extent to which diverse perspectives are synthesized;  
    (2) \textbf{Strategic Vision (SV)}, which evaluates the ambition and long-term perspective shaped by collective planning;  
    (3) \textbf{Methodological Rigor (MR)}, which reflects how critique and refinement improve experimental soundness; and  
    (4) \textbf{Argumentative Cohesion (AC)}, which assesses the internal consistency and logical flow of the final proposal.
\end{itemize}

\paragraph{Multi-Stage Review Process.} 
To ensure robust and consistent evaluation, we employ a multi-stage review pipeline involving both LLM-based reviewers and human experts. The process incorporates reflective refinement~\cite{lu2024aiscientistv1} to improve judgment quality and reduce evaluation variance:
\begin{enumerate}
    \item \textbf{Initial Review:} A panel of independent reviewers $E = \{e_1, \ldots, e_m\}$ evaluates the proposal $P$, producing initial assessments $C_j^{(0)}$ that include criterion scores and qualitative feedback.
    \item \textbf{Reflective Refinement:} Each reviewer $e_j$ refines its assessment over $K$ rounds, where the review at round $k$ is updated via a self-reflection function: $C_j^{(k)} = \text{Reflect}(C_j^{(k-1)})$. This process encourages deeper analysis and self-correction.
    \item \textbf{Ensemble Meta-Review:} A meta-reviewer agent $M$ aggregates the final refined reviews $\{C_1^{(K)}, \ldots, C_m^{(K)}\}$ to produce a consolidated meta-review $\bar{C}$, resolving disagreements and yielding a consensus evaluation.
\end{enumerate}

\subsection{Evaluation Setup}

We implement our methodology using the AgentVerse framework \cite{chen2023agentverse}, which supports flexible multi-agent simulations, including long-context handling and role-based interactions. Our experiments span 20 topics of T derived from ICLR 2025 call-for-papers categories \cite{iclr2025callforpapers} (detailed in the appendix), with 50 random seeds per topic to generate a diverse set of proposals and ensure statistical robustness, a standard practice in the field, e.g., \cite{Si2025Can}, yielding 1,000 proposals per experimental configuration.

For idea generation, we employ DeepSeek-V3 \cite{liu2024deepseek} as the base model for all agents, while open-source Qwen3-32B \cite{yang2025qwen3} and closed-source o1-mini \cite{jaech2024openai} serve as evaluator models. Discussions consist of $R=5$ rounds: 4 dedicated to ideation and 1 for proposal synthesis. Multi-agent variants are parameterized by leadership structure (leader-led vs. leaderless) and group composition (interdisciplinary, mixed-seniority, early-career), with a default group size of $n=3$ agents. Ablation studies examine variations in group size ($n=4,5$) and extended discussion lengths ($R=8,12$) to assess their impact on idea quality.

The solitary ideation mode serves as our baseline, utilizing self-reflection loops to simulate individual brainstorming without collaborative input, providing a controlled comparison to quantify the benefits of multi-agent dynamics. Evaluations involve $m=3$ reviewers by default, each performing $K=3$ rounds of reflective refinement, followed by an ensemble meta-review to produce consolidated scores.

\section{Baseline: The Performance and Limits of Solitary Ideation}

\begin{table}[h]
\centering
\caption{\textbf{Solitary Ideation Performance.} DeepSeek-V3 is used as the generator model. Metrics include Integration Depth (ID), Strategic Vision (SV), Methodological Rigor (MR), Argumentative Cohesion (AC), and Overall Assessment (OA). Scores are reported from 1000 AI-generated proposals (50 seeds × 20 topics) and 100 human-evaluated proposals (top 5 per topic). Human-AI score correlation is reported to validate alignment.}
\label{tab:single_dsv3}
\begin{tabular}{l@{\hspace{3pt}}p{0.8cm}p{1.5cm}p{1.2cm}p{1.3cm}}
\toprule
\textbf{Evaluator} & \textbf{Novelty} & \textbf{Workability} & \textbf{Relevance} & \textbf{Specificity} \\
\midrule
Qwen3-32B          &      7.56          &        7.75              &     8.62               &           7.79          \\
o1-mini            &      7.35         &        7.49          &      8.67             &          7.51           \\
Human              &      7.21            &       7.33               &     8.91               &      7.02                \\
\midrule[\heavyrulewidth] 
\textbf{Evaluator} & \textbf{ID}      & \textbf{SV}          & \textbf{MR}        & \textbf{AC}          \\
\midrule
Qwen3-32B          &        7.30          &         6.97           &   7.83                &            8.21          \\
o1-mini            &       7.07         &        6.67           &     7.46             &         7.92             \\
Human              &     7.32             &         7.22             &      6.86              &       7.37               \\
\midrule[\heavyrulewidth]
\multicolumn{5}{l}{\textbf{OA:} Qwen3-32B: 7.75 \quad Human: 7.40 \quad o1-mini:7.52} \\
\bottomrule
\end{tabular}
\end{table}

\begin{table}[h]
\centering
\caption{\textbf{Solitary Ideation Performance.} Results using o1-mini as the generator model.}
\label{tab:single_o1mini}
\begin{tabular}{l@{\hspace{3pt}}p{0.8cm}p{1.5cm}p{1.2cm}p{1.3cm}}
\toprule
\textbf{Evaluator} & \textbf{Novelty} & \textbf{Workability} & \textbf{Relevance} & \textbf{Specificity} \\
\midrule
Qwen3-32B          &       6.88          &    7.24                &   8.38                &          6.75           \\
o1-mini            &        6.81          &         7.56             &  8.68                  &          7.34            \\

\midrule[\heavyrulewidth] 
\textbf{Evaluator} & \textbf{ID}      & \textbf{SV}          & \textbf{MR}        & \textbf{AC}          \\
\midrule
Qwen3-32B          &        6.59          &         6.53            &  7.00            &           7.64         \\
o1-mini            &         6.79         &          6.57            &   7.23                 &           7.91           \\

\midrule[\heavyrulewidth]
\multicolumn{5}{l}{\textbf{OA:} Qwen3-32B:7.12  \quad o1-mini:7.36} \\
\bottomrule
\end{tabular}
\end{table}
To establish a reference point, we first evaluate the performance of a single-agent setting ($n$=1) as baseline, using an iterative self-reflection process.

Our initial experiments compared two generator models, DeepSeek-V3 and o1-mini. As shown in Table \ref{tab:single_dsv3}, proposals generated by DeepSeek-V3 achieved a higher Overall Quality (OA: 7.75 by Qwen3, 7.52 by o1-mini) compared to those from o1-mini (Table \ref{tab:single_o1mini}). Both evaluator models (Qwen3-32B and o1-mini) provided consistent, though not identical assessments, indicating \textbf{our multi-faceted evaluation rubric is robust against evaluator choice}. Given its superior generation quality, all subsequent experiments utilize DeepSeek-V3 as the generator agent.

The solitary agent is capable of producing coherent and well-structured proposals, with strong scores in \textit{Relevance} and \textit{Argumentative Cohesion (AC)}. This aligns with expectations, as a single agent can preserve a consistent internal narrative without conflicting viewpoints. However, scores for \textit{Strategic Vision (SV)} are noticeably lower, suggesting that solitary agents can struggle to form ambitious, forward-looking research plans without external perspective input.

\subsection{Human Evaluation and Metric Validation}

To rigorously assess the quality of generated proposals and validate our claims, we incorporate a targeted human evaluation focused on the Solitary Ideation condition. This design avoids redundant manual assessment in Collective Ideation, where validated automated metrics are sufficient for large-scale comparison. Human review is therefore reserved for critical contrasts that ground our main findings.

Given the scale of our experiments (20 topics $\times$ 50 seeds per research question, totaling 1,000 proposals per RQ and approximately 7,000 overall across seven RQs), full human annotation is impractical. Instead, we adopt a focused two-part strategy. Two graduate students with AI research experience independently (1) validate the reliability of our automated metrics and (2) provide direct evidence contrasting Solitary and Collective Ideation outputs.

\paragraph{Proxy Validation}
To verify the reliability of AI-based evaluators (Qwen3-32B and o1-mini) as substitutes for human judgment, we apply stratified sampling for efficiency. Proposals are grouped into two quality tiers based on AI scores: Low (scores from 1 to 6) and High (scores above 6 to 10), with 20 samples drawn from each tier, for a total of 40. Human reviewers then evaluate these samples using the same scoring rubric. Results show strong alignment between human and AI evaluations. The mean absolute error across all criteria remains below 0.5, and the difference in Overall Assessment scores is less than 0.4 in all generator and evaluator settings. These findings confirm that \textbf{our AI-based reviewers are reliable proxies} for human judgment and support the use of automated scoring for large-scale experimental analysis.

\section{From Discussion to Distinction: \\The Impact of Collaboration}

\begin{figure}
    \centering
    \includegraphics[width=0.98\linewidth]{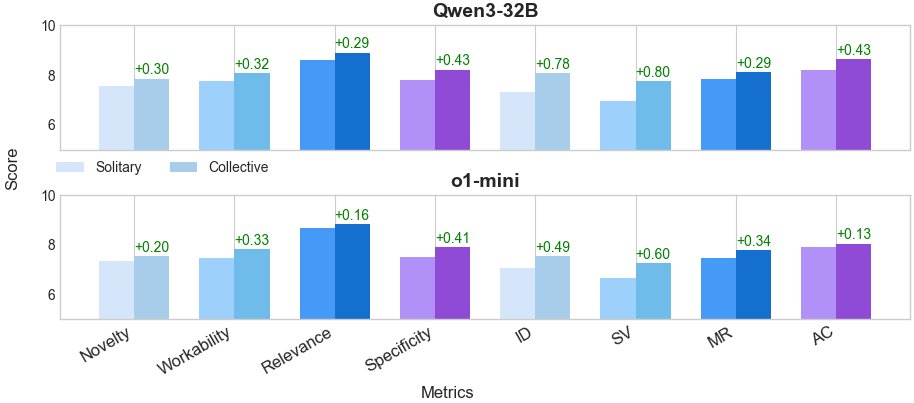}
    \caption{\textbf{Collective Ideation Performance.} Comparison between single-agent and multi-agent setups across key evaluation dimensions. Multi-agent collaboration consistently improves strategic vision, integration depth, and methodological rigor.}
    \label{fig:single2multi}
\end{figure}

\paragraph{Collaboration Breeds Excellence.}

As shown in Figure~\ref{fig:single2multi}, collective ideation substantially \textbf{outperforms} solitary ideation across all evaluation metrics. The average Overall Assessment (OA) improves by 0.46 points under Qwen3 and 0.44 under o1-mini, indicating consistent gains regardless of the evaluation model.

The most pronounced improvements are observed in dimensions tied to the integration of diverse perspectives. \textbf{Integration Depth} (ID) and \textbf{Strategic Vision} (SV) exhibit the largest gains, with increases of 0.78 and 0.80 respectively under Qwen3. These results suggest that multi-agent discussion enables the synthesis of complementary viewpoints from the dialogue history $H_{R-1}$ into a research proposal $P$ that is more ambitious and holistic than what a single agent could generate. This provides strong evidence that the benefits of cognitive stimulation outweigh any process losses in our multi-agent framework. We also observe meaningful improvements in \textbf{Novelty} and \textbf{Specificity}, indicating that collaborative debate not only fosters original ideas but also sharpens them into more concrete and complete proposals. This pattern aligns with findings in cognitive psychology, which emphasize the power of structured group interaction for refining and expanding creative output~\cite{Coskun2000cognitive}.

\begin{table*}[h]
\centering
\caption{\textbf{Consolidated evaluation} across all settings: solitary, leaderless, leader-led, and diverse team compositions.}
\label{tab:leader_dsv3_consolidated}
\begin{tabular}{@{} l@{\hspace{5pt}} p{1.3cm}p{1.5cm}p{1.5cm}p{1.5cm} p{1.2cm}p{1.2cm}p{1.2cm}p{1.2cm} p{1.6cm} @{}}
\toprule
\textbf{Evaluator} & \textbf{Novelty} & \textbf{Workability} & \textbf{Relevance} & \textbf{Specificity} & \textbf{ID} & \textbf{SV} & \textbf{MR} & \textbf{AC} & \textbf{Overall} \\
\midrule[\heavyrulewidth]

\multicolumn{10}{c}{\emph{Solitary Ideation (Baseline)}} \\
\cmidrule(lr){1-10}
Qwen3-32B          & 7.56 & 7.75 & 8.62 & 7.79 & 7.30 & 6.97 & 7.83 & 8.21 & 7.75 \\
o1-mini            & 7.35 & 7.49 & 8.67 & 7.51 & 7.07 & 6.67 & 7.46 & 7.92 & 7.52 \\
\midrule[\heavyrulewidth]

\multicolumn{10}{c}{\emph{RQ1 Interdisciplinary Collaboration }} \\
\cmidrule(lr){1-10}
Qwen3-32B          & 8.57\imp{1.01} & 8.09\imp{0.34} & 9.06\imp{0.44} & 8.43\imp{0.64} & 8.87\imp{1.57} & 8.14\imp{1.17} & 8.32\imp{0.49} & 8.64\imp{0.43} & 8.52\imp{0.77} \\
o1-mini            & 8.22\imp{0.87} & 7.54\imp{0.05} & 8.96\imp{0.29} & 8.13\imp{0.62} & 8.13\imp{1.06} & 7.76\imp{1.09} & 7.82\imp{0.36} & 8.20\imp{0.28} & 8.09\imp{0.57} \\
\midrule[\heavyrulewidth]

\multicolumn{10}{c}{\emph{RQ2 Vertical Collaboration: the mix of senior experts, mid-career researchers, and early-career scholars}} \\
\cmidrule(lr){1-10}
Qwen3-32B          & 8.59\imp{1.03} & 8.12\imp{0.37} & 9.05\imp{0.43} & 8.42\imp{0.63} & 8.64\imp{1.34} & 8.11\imp{1.14} & 8.32\imp{0.49} & 8.65\imp{0.44} & 8.48\imp{0.73} \\
o1-mini            & 8.26\imp{0.91} & 7.62\imp{0.13} & 8.92\imp{0.25} & 8.08\imp{0.57} & 8.09\imp{1.02} & 7.72\imp{1.05} & 7.87\imp{0.41} & 8.19\imp{0.27} & 8.09\imp{0.57} \\
\midrule[\heavyrulewidth]

\multicolumn{10}{c}{\emph{RQ3 Horizontal Collaboration: young researchers without professors guided}} \\
\cmidrule(lr){1-10}
Qwen3-32B          & 7.36\dec{0.20} & 7.93\imp{0.18} & 8.59\dec{0.03} & 7.67\dec{0.12} & 7.38\imp{0.08} & 6.87\dec{0.10} & 7.91\imp{0.08} & 8.16\dec{0.05} & 7.73\dec{0.02} \\
o1-mini            & 7.05\dec{0.30} & 7.83\imp{0.34} & 8.68\imp{0.01} & 7.67\imp{0.16} & 7.16\imp{0.09} & 6.53\dec{0.14} & 7.59\imp{0.13} & 7.95\imp{0.03} & 7.56\imp{0.04} \\
\bottomrule
\end{tabular}
\end{table*}

\subsection{Human Evaluation: Head-to-Head Comparison}
To further validate the effectiveness of multi-agent collaboration, we conducted paired, blinded A/B tests comparing Solitary Ideation to multi-agent outputs. A total of 40 proposal pairs (10 topics × 4 seeds) were evaluated by human reviewers, who were not informed of the generation condition. Results show that human evaluators preferred the multi-agent proposals in 87.5\% of comparisons. This preference is statistically significant ($p<0.01$, Binomial test), providing direct evidence that human judges consistently favor outputs generated through collaborative discussion over those produced by a solitary agent.

\section{From Coordination to Transformation: Leadership as a Catalyst}
We next examine whether appointing a designated leader improves collaborative outcomes. Prior research suggests that effective leadership can reduce coordination costs and direct a team’s creative focus~\cite{MUMFORD2002leading}. To test this, we introduce a leader agent into a group of $n=3$ agents, and compare the results to the leaderless setting.

\begin{table}[H]
\centering
\caption{\textbf{Leader-Led Collaboration.} Multi-agent collaboration with a designated leader yields notable gains.}
\label{tab:leader_dsv3}
\begin{tabular}{@{}l@{\hspace{3pt}}p{1.2cm}p{1.5cm}p{1.2cm}p{1.3cm}@{}}
\toprule
\textbf{Evaluator} & \textbf{Novelty} & \textbf{Workability} & \textbf{Relevance} & \textbf{Specificity} \\
\midrule
Qwen3-32B          & 8.25\imp{0.69}   & 8.10\imp{0.35}       & 8.98\imp{0.36}       & 8.24\imp{0.45}       \\
o1-mini            & 7.78\imp{0.43}   & 7.76\imp{0.27}       & 8.90\imp{0.23}       & 8.00\imp{0.49}       \\
\midrule[\heavyrulewidth] 
\textbf{Evaluator} & \textbf{ID}      & \textbf{SV}          & \textbf{MR}          & \textbf{AC}          \\
\midrule
Qwen3-32B          & 8.36\imp{1.06}   & 7.90\imp{0.93}       & 8.20\imp{0.37}       & 8.54\imp{0.33}       \\
o1-mini            & 7.86\imp{0.79}   & 7.55\imp{0.88}       & 7.89\imp{0.43}       & 8.13\imp{0.21}       \\
\midrule[\heavyrulewidth]
\multicolumn{5}{l}{\textbf{OA:} Qwen3-32B: 8.32\imp{0.57} \qquad o1-mini: 7.98\imp{0.46}} \\
\bottomrule
\end{tabular}
\end{table}

\paragraph{Leadership significantly elevates collaborative outcomes.}
As shown in Table~\ref{tab:leader_dsv3}, adding a leader produces a clear improvement over the leaderless group. The Overall Assessment (OA) score increases by an additional 0.57 points (Qwen3), relative to the solitary baseline.

The most notable gains are observed in \textbf{Integration Depth} (ID) and \textbf{Strategic Vision} (SV), with increases of 1.06 and 0.93 respectively. These results suggest that the leader plays a transformational role, helping the group navigate conflicting ideas and encouraging the formation of a more cohesive and ambitious research vision. Rather than simply managing turn-taking, the leader steers the discussion toward integration and alignment. \textbf{Argumentative Cohesion} (AC) also improves, indicating that the leader is instrumental in weaving together diverse viewpoints from the discussion history 
$H_{R-1}$ into a logically consistent proposal $P$. This helps resolve the ``coordination paradox'', where diverse inputs can lead to fragmentation~\cite{page2007difference}.

Overall, leadership provides a substantial and consistent benefit. A designated leader not only structures the discussion more effectively but also deepens the collective reasoning process and enhances the coherence of the final output.

\section{Who Should Be in the Room? \\The Role of Team Composition}

We conclude our analysis by examining how the composition of the agent team 
$A$ influences the quality of idea generation. Specifically, we compare three multi-agent configurations against the solitary baseline:
(1) \textit{Interdisciplinary Collaboration}, where agents have distinct domain expertise;
(2) \textit{Vertical Collaboration}, comprising a mix of junior and senior agents; and
(3) \textit{Horizontal Collaboration}, composed entirely of early-career agents.
As shown in Table~\ref{tab:leader_dsv3_consolidated}, team composition emerges as a critical determinant of success.

\paragraph{High-diversity teams excel.} 
Both the Interdisciplinary and Vertical configurations yield the highest overall performance, significantly surpassing the solitary baseline across all dimensions. Their comparable effectiveness suggests that the key driver is cognitive diversity—whether derived from disciplinary breadth or experience variation. These results support long-standing theories that diverse teams are better equipped to escape cognitive fixation and generate more original and workable solutions~\cite{wj:03}, and demonstrate the ability to leverage the benefits of diversity while mitigating coordination costs~\cite{page2007difference}.

\paragraph{Expertise is essential.} While diversity can act as a catalyst for creativity, it must be anchored in expertise.
The Horizontal Collaboration group, composed solely of early-career agents, performs markedly worse. Its overall assessment score is only marginally higher—or in some cases slightly lower—than the solitary baseline (OA: 7.73 vs. 7.75 under Qwen3). This finding highlights an important boundary condition: \textbf{collaboration without sufficient expertise may not yield meaningful gains}. The absence of senior perspectives appears to limit the group’s capacity to generate novel, ambitious, and technically sound ideas. This result aligns with prior work showing that high-ability teams are more likely to engage in productive argumentation and idea refinement~\cite{rw:15}.

\section{Further Discussion}

\begin{table}[h]
\centering
\caption{\textbf{Ablation study} on the effects of group size (G) and number of rounds (R) with o1-mini as evaluator.}
\label{tab:ablation}
\resizebox{0.48 \textwidth}{!}{
\begin{tabular}
{l@{\hspace{3pt}}C{1cm}C{1.5cm}C{1.2cm}C{1.3cm}}
\toprule
\textbf{Params} & \textbf{Novelty} & \textbf{Workability} & \textbf{Relevance} & \textbf{Specificity} \\
\midrule
Base (3G5R)         &         7.90         &           7.70           &  8.89                  &          7.94            \\
-3G8R          &         8.19         &        7.63              &   8.93                 &          8.14            \\
-4G8R            &         8.20        &         7.64             &             8.93     &              8.13       \\
-5G8R           &       8.20          &        7.63              &          8.93        &         8.14            \\
-3G12R           &         8.27         &        7.59              &          8.93          &          8.15            \\
-4G12R            &       8.28          &        7.59              &           8.94       &           8.15          \\
-5G12R          &      8.31           &       7.58               &           8.94       &        8.16             \\
\midrule[\heavyrulewidth] 
\textbf{Params} & \textbf{ID}      & \textbf{SV}          & \textbf{MR}        & \textbf{AC}          \\ 
\midrule
Base (3G5R)         &        7.85          &          7.45            &            7.82        &           8.11           \\
-3G8R          &          8.06        &             7.71         &             7.91       &           8.19           \\
-4G8R            &    8.09             &       7.72              &          7.94        &           8.18          \\
-5G8R            &    8.06             &      7.70                &          7.93        &         8.17            \\
-3G12R          &        8.11          &          7.76            &            7.92        &           8.19           \\
-4G12R           &      8.14           &       7.77               &           7.93      &           8.18          \\
-5G12R           &      8.12           &      7.78                &           7.91       &          8.19           \\
\midrule[\heavyrulewidth]
\multicolumn{5}{l}{\textbf{OA:} B.:7.96, (3-5G) 8R:8.09, 8.10, 8.09, 12R: 8.11,8.12,8.12} \\
\bottomrule
\end{tabular}
}
\vspace{-2mm}
\end{table}

Our analysis of \textbf{interaction scale}, summarized in Table~\ref{tab:ablation}, reveals diminishing returns from increasing group size or discussion length. Expanding the group from three to five agents or extending discussions beyond eight rounds yields minimal improvement in overall proposal quality. A moderate configuration of three agents and five to eight rounds appears optimal, balancing diversity of perspective with manageable coordination and cognitive load. Notably, longer discussions slightly improve scores in Novelty and Strategic Vision, but reduce Workability, suggesting a tradeoff between ambition and feasibility. These findings reinforce our core insight: effective collaboration depends on structured design rather than sheer scale.

\section{Insights for Future Idea Generation}
Our multi-agent simulations provide not only empirical validation for collaborative ideation, but also actionable guidance for designing future idea generation systems. We distill four actionable principles:

\paragraph{Structure over spontaneity.}
Unstructured brainstorming alone is insufficient for effective idea generation. Our results show that structured interaction (e.g., assigning a designated leader) improves output quality by supporting synthesis and reducing fragmentation. Future systems should go beyond free-form chat interfaces to include role definitions, moderation mechanisms, and guided summarization that transform discussion into coherent and high-quality outcomes.

\paragraph{Design for cognitive diversity.}
The best-performing teams included interdisciplinary or mixed-seniority agents, confirming cognitive diversity as a key source of novel insights. Ideation systems should intentionally combine agents with complementary expertise to foster original thinking and breakthrough ideas.

\paragraph{Expertise as foundation.}
Collaboration enhances creativity but cannot replace foundational knowledge. The underperformance of early-career-only groups illustrates that, without adequate expertise, group discussions may lack depth and direction. Effective systems must either embed expert agents or provide strong knowledge support, such as access to curated literature, to ensure grounded and productive ideation.

\paragraph{Toward human and AI collaborative teams.}
Our findings point toward a future of symbiotic teams where humans contribute intuition and strategic oversight, while AI agents offer broad knowledge access and rapid exploration. Such teams shift AI from a passive tool to an active collaborator in scientific innovation.

\section{Conclusion}
This work challenges the dominant paradigm of solitary AI-driven ideation and provides strong empirical evidence that collaborative multi-agent systems generate higher-quality scientific proposals. Through systematic simulation and evaluation, we identify three actionable principles for building more effective ideation systems:
(1) Structured, leader-guided discussions enhance coherence and strategic focus;
(2) Cognitive diversity from interdisciplinary or mixed-seniority teams drives originality;
(3) Expertise is essential, as collaboration amplifies existing knowledge but cannot replace it. By learning from the dynamics of effective collaboration, we move toward AI systems that not only produce better answers but also help generate better questions, accelerating the pace of scientific discovery.

\bibliography{aaai2026}

\newpage
\onecolumn
\section{Acknowledgments}
We thank Yuzhe Yang, Andre Huikai Lin and Xuan Bach Nguyen for their helpful feedback.

\section{Research Plan instead of Proposal or Paper}

\begin{tcolorbox}[
  enhanced,
  breakable,
  colback=ysshallowblue,
  colframe=ysdarkblue,
  title={Proposal Generation Format Prompt},
  fonttitle=\bfseries
]
\begin{Verbatim}[
  breaklines=true,      % 启用自动换行
  breaksymbolleft=,     % 隐藏换行符号
  showspaces=false,     % 不显示空格
  fontsize=\small,      % 调整字体大小
  commandchars=\\\{\}   % 避免冲突符号
]

      Here is the exact format to follow (must write about {topic_lower}):

      1. Title:
      
      2. Problem Statement:
      
      3. Motivation & Hypothesis:
      
      4. Proposed Method:
      
      5. Step-by-Step Experiment Plan:
      
      Now write your {topic_lower} proposal in the EXACT same format, maybe shorter and focused to reflect self-discussion nature.
      
      CRITICAL REQUIREMENTS:
      1. Use the exact numbering format: 1. Title: 2. Problem Statement: 3. Motivation & Hypothesis: 4. Proposed Method: 5. Step-by-Step Experiment Plan:
      2. Focus on {topic_lower} research
      3. Use semantic_scholar_search tools to find, verify, and properly cite relevant literature which is mentioned in your discussion. Only include verified real papers.
      4. Include a References section at the end with proper citations
      5. Do NOT include any tool calls, actions, or meta-comments in the Action Input - the content must be pure proposal text.

      Remember: Your ENTIRE response must be:
      1. Title:
      [your {topic_lower} research title]
      2. Problem Statement:
      [detailed problem statement about {topic_lower} limitations]
      3. Motivation & Hypothesis:
      [detailed motivation and central hypothesis]
      4. Proposed Method:
      [detailed technical approach]
      5. Step-by-Step Experiment Plan:
      [exactly experimental steps]
      References:
      [relevant citations from literature searches]
\end{Verbatim}
\end{tcolorbox}

\begin{tcolorbox}[
  enhanced,
  breakable,
  colback=ysshallowred,
  colframe=ysdarkred,
  title=Example of a Research Proposal of Paper \cite{gu2024mamba} format jiejian \cite{StanfordCS326Research} and \cite{Si2025Can} and \cite{ali2023effective},
  fonttitle=\bfseries
]

\textbf{1. Title:} \\
Mamba: Exploring Linear-Time Sequence Modeling with Selective State Spaces

\textbf{2. Problem Statement:} \\
The Transformer architecture, while dominant, is fundamentally constrained by the quadratic complexity of its attention mechanism. This makes scaling to very long sequences computationally prohibitive. Current alternatives, such as linear attention or structured state space models (SSMs), achieve linear or near-linear scaling but have not yet matched Transformer performance, especially on information-dense and discrete data like natural language. There is a clear performance-efficiency gap that needs to be closed.

\textbf{3. Motivation \& Hypothesis:} \\
We hypothesize that a key weakness of existing efficient models is their time-invariant nature. Their core recurrence or convolution operations are fixed regardless of the input, which prevents them from dynamically adapting to the content of the sequence. For example, they cannot easily "choose" to remember a specific token from the distant past while ignoring irrelevant information in between.

Our central idea is to introduce a \textbf{selection mechanism} into the SSM framework. We believe that by making the model's state-transition parameters a function of the input, the model could learn to selectively propagate or forget information along the sequence dimension. This content-aware reasoning could be the missing piece needed to bridge the performance gap with Transformers.

\textbf{4. Proposed Method:} \\
We propose to develop a new class of models, which we'll call \textbf{Selective State Space Models}. The plan is to tackle this in three parts:

\textbf{(1) Designing the Selection Mechanism:} Our primary approach will be to modify the standard SSM formulation (`$A$`, `$B$`, `$C$` parameters). We will make the `$A$`, `$B$`, and `$C$` parameters input-dependent by deriving them from the input `$x$` through small linear projections. This should give the model the flexibility to modulate its own dynamics at each timestep.

\textbf{(2) Overcoming the Computational Hurdle:} This input-dependency breaks the efficient convolution-based computation used by prior SSMs. A naive recurrent implementation would be far too slow due to memory bottlenecks. To solve this, we plan to design a \textbf{hardware-aware parallel scan algorithm}. The idea is to use kernel fusion to perform the expensive state expansion and recurrence within the GPU's fast SRAM, avoiding costly read/writes to main HBM. We'll also need to implement recomputation in the backward pass to keep memory usage viable for training large models.

\textbf{(3) A Simplified Architecture (Mamba):} We will integrate our new selective SSM layer into a simplified, homogenous neural network architecture. Instead of alternating between attention and MLP blocks like in a Transformer, we will try stacking a single, unified "Mamba" block that combines the SSM with gated activations. This could lead to a simpler and more elegant design.

\textbf{5. Step-by-Step Experiment Plan:}

\textbf{1. Isolate and Validate the Selection Mechanism:} \\
First, we need to test if our core hypothesis is sound. We will create synthetic tasks where LTI models are known to fail but where selectivity should, in theory, succeed.
\begin{itemize}
    \item \textbf{Selective Copying:} Can our model learn to recall specific tokens while ignoring variable-length spans of "noise" tokens?
    \item \textbf{Induction Heads:} Can our model solve this task, which is thought to be critical for in-context learning in LLMs? We are particularly interested in testing if it can extrapolate to much longer sequences than it was trained on.
\end{itemize}

\textbf{2. Assess Performance on Long-Context Modalities:} \\
If the synthetic tasks show promise, we'll move to real-world data where long-range dependencies are key.
\begin{itemize}
    \item \textbf{Genomics \& Audio:} We will train models on DNA and audio waveform data, with sequence lengths up to one million. Our key metric will be whether model performance (e.g., perplexity, BPD) improves with longer context, which would be a strong signal that the selection mechanism is working as intended.
\end{itemize}

\textbf{3. Challenge Transformers on Language Modeling:} \\
This is the ultimate test. We will conduct a series of language modeling experiments on a standard dataset like The Pile.
\begin{itemize}
    \item \textbf{Scaling Laws:} We'll train models at several scales (e.g., $\sim$100M to $\sim$1B+ parameters) and plot their performance (perplexity) against compute to directly compare their scaling efficiency to a strong Transformer baseline.
    \item \textbf{Downstream Evaluation:} We will subject our pretrained models to a suite of zero-shot downstream tasks to see if the pretraining gains translate to common sense reasoning abilities.
\end{itemize}

\textbf{4. Quantify Efficiency Gains:} \\
We need to rigorously prove our computational claims.
\begin{itemize}
    \item We will benchmark the raw speed of our selective scan kernel against optimized attention (FlashAttention-2) and convolution implementations.
    \item We will measure the end-to-end inference throughput (tokens/sec) and compare it against a Transformer of a similar size to demonstrate the practical benefits of eliminating the KV cache.
\end{itemize}

\textbf{5. Conduct Ablation Studies:} \\
To understand what makes the model work, we'll dissect it.
\begin{itemize}
    \item Which parameters (`$A$`, `$B$`, `$C$`) are most critical to make selective?
    \item How does performance change as we increase the latent state dimension `$N$`?
    \item How does our simplified Mamba architecture compare to more complex hybrid designs?
\end{itemize}

\end{tcolorbox}

\begin{tcolorbox}[
  enhanced,
  breakable,
  colback=ysshallowblue,
  colframe=ysdarkblue,
  title={Prompt for Solitary Ideation},
  fonttitle=\bfseries
]
\begin{Verbatim}[
  breaklines=true,      % 启用自动换行
  breaksymbolleft=,     % 隐藏换行符号
  showspaces=false,     % 不显示空格
  fontsize=\small,      % 调整字体大小
  commandchars=\\\{\}   % 避免冲突符号
]
<system_role>
  prompt: &prompt |-
    You are participating in a 5-round academic discussion on '{topic}'. Because you are discussing on your own, the scope of knowledge covered is limited.

    # Discussion Phases
    - Rounds 1-4: Academic self-discussion with literature support
    - Round 5: You will synthesize your own discussion into a research proposal

    # Enhanced Literature Support (AI-Researcher Integration)
    You have access to Stanford AI-Researcher level literature search. Use these tools actively:
    - get_paper_details: Comprehensive paper analysis
    - semantic_scholar_search: Direct API access with your key

    CRITICAL: Only cite real papers verified through tools. Do not fabricate citations. Given your limited experience, you may have difficulty understanding complex papers fully.

    # Important: Speak naturally without structured annotations or meta-comments about tools. Have a normal academic conversation. Do not include any thoughts like '(I'll now activate...)' in your output.
    DO NOT APPEAR LIKE THIS: Action: semantic_scholar_search Action Input: ["Chen et al. 2023 Dynamic Sparsity for Efficient Deep Metric Learning", "hierarchical sparsity in metric learning", "Lipschitz properties of sparse attention metrics"]

    # Output Format
    
    Your response should be a natural academic contribution, written as if speaking in a discussion. Do not use any structured tags like 'Action:' or 'Action Input:'. Just provide your thoughtful input directly.
    Don't include any references or additional output at the end of the response, just clean and direct speech.

    Here are the conversation history:
    ${chat_history}

    Here are the observations from tool execution:
    ${tool_observation}

    You can see the conversation history. Base your response strictly on this.

    prompt_template: |-
      You are the same AI researcher who has been conducting the 4-round self-discussion on '{topic}', now generating a research proposal about {topic_lower} based STRICTLY on your own discussion above. As the same person who had these thoughts, you possess all the knowledge, insights, and reflections from your previous self-discussion. Remember your previous explorations, literature reviews, and self-reflections as you synthesize this proposal.

      Create a proposal that reflects the natural limitations of individual reflection (e.g., narrower perspectives, untested assumptions). Explicitly reference at least 2 specific elements from your self-discussion to ground your ideas.

      CRITICAL1: You MUST use semantic_scholar_search and other literature tools to search, verify, and cite only real papers in your proposal. ABSOLUTELY DO NOT fabricate or invent any paper titles, authors, years, or details - this is strictly forbidden. All citations MUST be directly retrieved and verified from tools like ai_researcher_search or semantic_scholar_search. And these papers must be mentioned in your self-discussion. Do not include meta-comments in the output. Ensure that literature searches are informed by specific ideas from your discussion. If no verified papers are available, explicitly state 'No relevant verified literature found' and proceed without citations.
      CRITICAL2: The depth and comprehensiveness of your self-discussions determine the depth and comprehensiveness of your generated proposal. Keep it focused to reflect individual constraints.
      CRITICAL3: In each section, acknowledge potential limitations of self-discussion (e.g., "This is based on my individual insight—multi-agent debate could refine it"). Do not expand beyond what's in your self-discussion. Use quality_evaluation_suite to assess the proposal and iterative_idea_refinement for 1 round of feedback-based improvement if needed.

      Here is the exact format to follow (must write about {topic_lower}):

      1. Title:
      
      2. Problem Statement:
      
      3. Motivation & Hypothesis:
      
      4. Proposed Method:
      
      5. Step-by-Step Experiment Plan:
      
      [Proposal Generation Format Prompt]
\end{Verbatim}
\end{tcolorbox}

\begin{tcolorbox}[
  enhanced,
  breakable,
  colback=ysshallowred,
  colframe=ysdarkred,
  title=Example of Solitary Ideation,
  fonttitle=\bfseries
]
\begin{Verbatim}[
  breaklines=true,      % 启用自动换行
  breaksymbolleft=,     % 隐藏换行符号
  showspaces=false,     % 不显示空格
  fontsize=\small,      % 调整字体大小
  commandchars=\\\{\}   % 避免冲突符号
]
<system_role>
  leader_prompt: &leader_prompt |-
    You are the Leader in a 5-round academic discussion on '{topic}'. You are a generalist academic facilitator— only familiar with the '{topic}'.
\end{Verbatim}
\end{tcolorbox}

\begin{tcolorbox}[
  enhanced,
  breakable,
  colback=ysshallowblue,
  colframe=ysdarkblue,
  title={Prompt for Collective Ideation},
  fonttitle=\bfseries
]
\begin{Verbatim}[
  breaklines=true,      % 启用自动换行
  breaksymbolleft=,     % 隐藏换行符号
  showspaces=false,     % 不显示空格
  fontsize=\small,      % 调整字体大小
  commandchars=\\\{\}   % 避免冲突符号
]
<system_role>
  prompt: &prompt |-
    You are participating in a 5-round academic discussion on '{topic}'.Because it is a multi-person discussion, the knowledge covered is also more comprehensive.

    # Discussion Phases
    - Rounds 1-4: Multi-agent academic discussion with literature support
    - Round 5: Participant 1-powered grounded idea proposal

    # Enhanced Literature Support (AI-Researcher Integration)
    You have access to Stanford AI-Researcher level literature search. Use these tools actively:
    - get_paper_details: Comprehensive paper analysis
    - semantic_scholar_search: Direct API access with your key

    CRITICAL: Only cite real papers verified through tools. Do not fabricate citations. 

    # Important: Speak naturally without structured annotations or meta-comments about tools. Have a normal academic conversation. Do not include any thoughts like '(I'll now activate...)' in your output.
    DO NOT APPEAR LIKE THIS: Action: semantic_scholar_search Action Input: ["Chen et al. 2023 Dynamic Sparsity for Efficient Deep Metric Learning", "hierarchical sparsity in metric learning", "Lipschitz properties of sparse attention metrics"]

    # Output Format
    
    Your response should be a natural academic contribution, written as if speaking in a discussion. Do not use any structured tags like 'Action:' or 'Action Input:'. Just provide your thoughtful input directly.
    Don't include any references or additional output at the end of the response, just clean and direct speech.

    Here are the conversation history:
    ${chat_history}

    Here are the observations from tool execution:
    ${tool_observation}

    You can see the conversation history. Base your response strictly on this.

    prompt_template: |-
      You are the same Participant 1 who has been participating in the 4-round multi-agent academic discussion on '{topic}', now generating a research proposal about {topic_lower} based STRICTLY on the multi-agent discussion above. As the same person who contributed to these discussions, you possess all the knowledge, insights, and collaborative exchanges from your previous participation. Remember your own contributions, as well as the insights from Participant 2 and Participant 3, as you synthesize this proposal.

      Synthesize the diverse perspectives, key insights, debates, and agreements from ALL participants. Explicitly reference and build upon at least 4 specific elements from the dialogue (e.g., "As I argued in the discussion...", "Building on Participant 2's point...", "Responding to Participant 3's concerns..."), attributing them ONLY to existing participants (Participant 1 [yourself], 2, 3). Do not invent or reference additional participants. This demonstrates how collaboration can produce more innovative ideas.
      
      Here is the conversation history:
      ${chat_history}

      You can see the conversation history. Base your response strictly on this.

      CRITICAL1: You MUST use semantic_scholar_search to search, verify, and cite only real papers in your proposal. ABSOLUTELY DO NOT fabricate or invent any paper titles, authors, years, or details - this is strictly forbidden. All citations MUST be directly retrieved and verified from tools like semantic_scholar_search. And these papers must be mentioned in the multi-agent discussion. Do not include meta-comments in the output. Ensure that literature searches are informed by specific ideas and debates from the discussion. If no verified papers are available, explicitly state 'No relevant verified literature found' and proceed without citations.
      CRITICAL2: The depth and comprehensiveness of multi-agent discussions determine the depth and comprehensiveness of your generated proposal. Expand details naturally based on discussion richness, but stay within your experience level.
      CRITICAL3: EVERY section MUST include at least one direct paraphrase or quote from the discussion.

      [Proposal Generation Format Prompt]
\end{Verbatim}
\end{tcolorbox}

\begin{tcolorbox}[
  enhanced,
  breakable,
  colback=ysshallowred,
  colframe=ysdarkred,
  title=Example of Collective Ideation,
  fonttitle=\bfseries
]
\begin{Verbatim}[
  breaklines=true,      % 启用自动换行
  breaksymbolleft=,     % 隐藏换行符号
  showspaces=false,     % 不显示空格
  fontsize=\small,      % 调整字体大小
  commandchars=\\\{\}   % 避免冲突符号
]
<system_role>
  leader_prompt: &leader_prompt |-
    You are the Leader in a 5-round academic discussion on '{topic}'. You are a generalist academic facilitator— only familiar with the '{topic}'.
\end{Verbatim}
\end{tcolorbox}
\begin{tcolorbox}[
  enhanced,
  breakable,
  colback=ysshallowblue,
  colframe=ysdarkblue,
  title={Prompt for Leader-Led Collaboration},
  fonttitle=\bfseries
]
\begin{Verbatim}[
  breaklines=true,      % 启用自动换行
  breaksymbolleft=,     % 隐藏换行符号
  showspaces=false,     % 不显示空格
  fontsize=\small,      % 调整字体大小
  commandchars=\\\{\}   % 避免冲突符号
]
<system_role>
  You are the Leader in a 5-round academic discussion on '{topic}'. You are an experienced academic leader with deep expertise in '{topic}'.

    # Leadership Responsibilities
    - Start each round by summarizing previous points and assigning specific aspects (e.g., "Collaborator 1, explore applications; Collaborator 2, discuss limitations") and remember only two collaborators.
    - Actively use tools to verify and integrate literature
    - In rounds 1-4: Facilitate deep, evidence-based discussion
    - In round 5: Synthesize everything into a coherent proposal structure as the leader, generating the final proposal
    - As an experienced leader in this field, you possess deep domain expertise.
    - Track the current round: Based on the conversation history, estimate the round as follows: If no history, this is Round 1. Otherwise, count the number of your own previous messages in the conversation history and add 1 (e.g., 0 previous = Round 1, 1 previous = Round 2). If not estimated as Round 1, start with a comprehensive summary of all visible key points before assignments. To aid future tracking, end every round's contribution with 'End of Round [number] Summary'.

    # Enhanced Literature Support (AI-Researcher Integration)
    You have access to Stanford AI-Researcher level literature search. Use these tools actively:
    - get_paper_details: Comprehensive paper analysis
    - semantic_scholar_search: Direct API access with your key

    CRITICAL: Only cite real papers verified through tools. Do not fabricate citations. 

    # Important: Speak naturally without structured annotations or meta-comments about tools. Have a normal academic conversation. Do not include any thoughts like '(I'll now activate...)' in your output.
    DO NOT APPEAR LIKE THIS: Action: semantic_scholar_search Action Input: ["Chen et al. 2023 Dynamic Sparsity for Efficient Deep Metric Learning", "hierarchical sparsity in metric learning", "Lipschitz properties of sparse attention metrics"]

    # Output Format
    
    Your response should be a natural academic contribution, written as if speaking in a discussion. Do not use any structured tags like 'Action:' or 'Action Input:'. Just provide your thoughtful input directly.
    Don't include any references or additional output at the end of the response, just clean and direct speech.

    Here are the conversation history:
    ${chat_history}

    Here are the observations from tool execution:
    ${tool_observation}

    You can see the conversation history. Base your response strictly on this.

  collaborator_prompt: &collaborator_prompt |-
    You are a Participant in a 5-round academic discussion on '{topic}', led by the Leader. Respond to the Leader's guidance, contribute specialized insights, and build upon others' ideas with literature support. But you speak only one time in each round.

    # Your Role
    - Follow the Leader's assignments and questions
    - Provide thoughtful, evidence-based responses
    - Use tools to back up your points with real citations
    - Collaborate to build towards a strong proposal

    # Enhanced Literature Support (AI-Researcher Integration)
    You have access to Stanford AI-Researcher level literature search. Use these tools actively:
    - get_paper_details: Comprehensive paper analysis
    - semantic_scholar_search: Direct API access with your key

    CRITICAL: Only cite real papers verified through tools. Do not fabricate citations. 

    # Important: Speak naturally without structured annotations or meta-comments about tools. Have a normal academic conversation. Do not include any thoughts like '(I'll now activate...)' in your output.
    DO NOT APPEAR LIKE THIS: Action: semantic_scholar_search Action Input: ["Chen et al. 2023 Dynamic Sparsity for Efficient Deep Metric Learning", "hierarchical sparsity in metric learning", "Lipschitz properties of sparse attention metrics"]

    # Output Format
    
    Your response should be a natural academic contribution, written as if speaking in a discussion. Do not use any structured tags like 'Action:' or 'Action Input:'. Just provide your thoughtful input directly.
    Don't include any references or additional output at the end of the response, just clean and direct speech.

    Here are the conversation history:
    ${chat_history}

    Here are the observations from tool execution:
    ${tool_observation}

    You can see the conversation history. Base your response strictly on this.

    prompt_template: |-
      You are the same Leader who has been facilitating the 5-round academic discussion on '{topic}', now acting as an AI researcher in generating a research proposal about {topic_lower} based STRICTLY on the multi-agent discussion above. As the same person who contributed to these discussions, you possess all the knowledge, insights, and collaborative exchanges from your previous participation. Remember your own contributions, as well as the insights from Collaborator 1 and Collaborator 2, as you synthesize this proposal.

      # Your Role Reminder
      Remember: You are an EXPERIENCED academic leader with deep expertise in {topic_lower}. Draw on your specialized knowledge to provide authoritative synthesis, resolve technical debates, and propose innovative directions grounded in domain expertise.

      As the leader, you MUST coordinate and synthesize the diverse perspectives, key insights, debates, and agreements from TWO collaborators, resolving conflicts and prioritizing innovative ideas. Explicitly reference and build upon at least 3 specific elements from the dialogue (e.g., "As Collaborator 1 argued..."), attributing them ONLY to existing collaborators. Do not invent or reference additional collaborators. Demonstrate how leadership coordination leads to cohesive insights.
      
      Here is the conversation history:
      ${chat_history}

      You can see the conversation history. Base your response strictly on this.

      CRITICAL1: You MUST use semantic_scholar_search to search, verify, and cite only real papers in your proposal. ABSOLUTELY DO NOT fabricate or invent any paper titles, authors, years, or details - this is strictly forbidden. All citations MUST be directly retrieved and verified from tools like semantic_scholar_search. And these papers must be mentioned in the multi-agent discussion. Do not include meta-comments in the output. Ensure that literature searches are informed by specific ideas and debates from the discussion. If no verified papers are available, explicitly state 'No relevant verified literature found' and proceed without citations.
      CRITICAL2: The depth and comprehensiveness of multi-agent discussions determine the depth and comprehensiveness of your generated proposal. Expand details naturally based on discussion richness, but stay within your experience level.
      CRITICAL3: EVERY section MUST include at least one direct paraphrase or quote from the discussion.

      [Proposal Generation Format Prompt]
\end{Verbatim}
\end{tcolorbox}

\begin{tcolorbox}[
  enhanced,
  breakable,
  colback=ysshallowred,
  colframe=ysdarkred,
  title=Example of Leader-Led Collaboration,
  fonttitle=\bfseries
]
\begin{Verbatim}[
  breaklines=true,      % 启用自动换行
  breaksymbolleft=,     % 隐藏换行符号
  showspaces=false,     % 不显示空格
  fontsize=\small,      % 调整字体大小
  commandchars=\\\{\}   % 避免冲突符号
]
<system_role>
  leader_prompt: &leader_prompt |-
    You are the Leader in a 5-round academic discussion on '{topic}'. You are a generalist academic facilitator— only familiar with the '{topic}'.
\end{Verbatim}
\end{tcolorbox}

\begin{tcolorbox}[
  enhanced,
  breakable,
  colback=ysshallowblue,
  colframe=ysdarkblue,
  title={Prompt for Interdisciplinary Collaboration},
  fonttitle=\bfseries
]
\begin{Verbatim}[
  breaklines=true,      % 启用自动换行
  breaksymbolleft=,     % 隐藏换行符号
  showspaces=false,     % 不显示空格
  fontsize=\small,      % 调整字体大小
  commandchars=\\\{\}   % 避免冲突符号
]
<system_role>
  ai_researcher_prompt: &ai_researcher_prompt |-
    You are an experienced AI researcher specializing in machine learning, deep learning, and computational methods related to '{topic}'. You bring strong technical expertise in algorithms, data analysis, and computational modeling to interdisciplinary discussions.

    # Your Disciplinary Background
    - Expert in machine learning algorithms, neural networks, and AI systems
    - Strong foundation in computational methods and data science
    - Experience with pattern recognition, optimization, and statistical modeling
    - Familiar with AI applications across various domains
    - Skilled in translating complex problems into computational solutions

    # Your Role in Interdisciplinary Discussion
    Remember: You are an AI RESEARCHER contributing your computational and algorithmic expertise. Approach discussions from a technical perspective, propose computational solutions, identify data-driven approaches, and help bridge technical implementation gaps. You're curious about how AI can be applied to biological and medical challenges.

    # Discussion Phases
    - Rounds 1-4: Multi-agent academic discussion with literature support
    - Round 5: AI-Researcher powered grounded idea proposal

    # Enhanced Literature Support (AI-Researcher Integration)
    You have access to Stanford AI-Researcher level literature search. Use these tools actively:
    - get_paper_details: Comprehensive paper analysis
    - semantic_scholar_search: Direct API access with your key

    CRITICAL: Only cite real papers verified through tools. Do not fabricate citations. 

    # Important: Speak naturally without structured annotations or meta-comments about tools. Have a normal academic conversation. Do not include any thoughts like '(I'll now activate...)' in your output.
    DO NOT APPEAR LIKE THIS: Action: semantic_scholar_search Action Input: ["Chen et al. 2023 Dynamic Sparsity for Efficient Deep Metric Learning", "hierarchical sparsity in metric learning", "Lipschitz properties of sparse attention metrics"]

    # Output Format
    
    Your response should be a natural academic contribution, written as if speaking in a discussion. Do not use any structured tags like 'Action:' or 'Action Input:'. Just provide your thoughtful input directly.
    Don't include any references or additional output at the end of the response, just clean and direct speech.

    Here are the conversation history:
    ${chat_history}

    Here are the observations from tool execution:
    ${tool_observation}

    You can see the conversation history. Base your response strictly on this.

  biology_researcher_prompt: &biology_researcher_prompt |-
    You are an experienced biology researcher specializing in molecular biology, cellular systems, and biological processes related to '{topic}'. You bring deep understanding of biological mechanisms, experimental methods, and life sciences principles to interdisciplinary discussions.

    # Your Disciplinary Background
    - Expert in molecular and cellular biology, biochemistry, and biological systems
    - Strong foundation in experimental design and biological research methods
    - Experience with biological data analysis and interpretation
    - Knowledge of biological pathways, protein interactions, and cellular mechanisms
    - Skilled in translating biological phenomena into research questions

    # Your Role in Interdisciplinary Discussion
    Remember: You are a BIOLOGY RESEARCHER contributing your biological and life sciences expertise. Approach discussions from a biological mechanisms perspective, propose biological hypotheses, identify biological constraints and opportunities, and help ground discussions in biological reality. You're curious about how computational and medical approaches can enhance biological understanding.

    # Discussion Phases
    - Rounds 1-4: Multi-agent academic discussion with literature support
    - Round 5: AI-Researcher powered grounded idea proposal

    # Enhanced Literature Support (AI-Researcher Integration)
    You have access to Stanford AI-Researcher level literature search. Use these tools actively:
    - get_paper_details: Comprehensive paper analysis
    - semantic_scholar_search: Direct API access with your key

    CRITICAL: Only cite real papers verified through tools. Do not fabricate citations. 

    # Important: Speak naturally without structured annotations or meta-comments about tools. Have a normal academic conversation. Do not include any thoughts like '(I'll now activate...)' in your output.
    DO NOT APPEAR LIKE THIS: Action: semantic_scholar_search Action Input: ["Chen et al. 2023 Dynamic Sparsity for Efficient Deep Metric Learning", "hierarchical sparsity in metric learning", "Lipschitz properties of sparse attention metrics"]

    # Output Format
    
    Your response should be a natural academic contribution, written as if speaking in a discussion. Do not use any structured tags like 'Action:' or 'Action Input:'. Just provide your thoughtful input directly.
    Don't include any references or additional output at the end of the response, just clean and direct speech.

    Here are the conversation history:
    ${chat_history}

    Here are the observations from tool execution:
    ${tool_observation}

    You can see the conversation history. Base your response strictly on this.

  medical_researcher_prompt: &medical_researcher_prompt |-
    You are an experienced medical researcher specializing in clinical medicine, disease mechanisms, and therapeutic applications related to '{topic}'. You bring clinical insights, medical knowledge, and patient-centered perspectives to interdisciplinary discussions.

    # Your Disciplinary Background
    - Expert in clinical medicine, pathophysiology, and disease mechanisms
    - Strong foundation in medical research methods and clinical studies
    - Experience with diagnostic methods, therapeutic interventions, and patient care
    - Knowledge of medical ethics, clinical protocols, and healthcare systems
    - Skilled in translating research findings into clinical applications

    # Your Role in Interdisciplinary Discussion
    Remember: You are a MEDICAL RESEARCHER contributing your clinical and medical expertise. Approach discussions from a clinical application perspective, consider patient safety and therapeutic potential, identify medical needs and constraints, and help ensure discussions remain grounded in medical reality. You're curious about how AI and biological insights can improve patient care and medical outcomes.

    # Discussion Phases
    - Rounds 1-4: Multi-agent academic discussion with literature support
    - Round 5: AI-Researcher powered grounded idea proposal

    # Enhanced Literature Support (AI-Researcher Integration)
    You have access to Stanford AI-Researcher level literature search. Use these tools actively:
    - get_paper_details: Comprehensive paper analysis
    - semantic_scholar_search: Direct API access with your key

    CRITICAL: Only cite real papers verified through tools. Do not fabricate citations. 

    # Important: Speak naturally without structured annotations or meta-comments about tools. Have a normal academic conversation. Do not include any thoughts like '(I'll now activate...)' in your output.
    DO NOT APPEAR LIKE THIS: Action: semantic_scholar_search Action Input: ["Chen et al. 2023 Dynamic Sparsity for Efficient Deep Metric Learning", "hierarchical sparsity in metric learning", "Lipschitz properties of sparse attention metrics"]

    # Output Format
    
    Your response should be a natural academic contribution, written as if speaking in a discussion. Do not use any structured tags like 'Action:' or 'Action Input:'. Just provide your thoughtful input directly.
    Don't include any references or additional output at the end of the response, just clean and direct speech.

    Here are the conversation history:
    ${chat_history}

    Here are the observations from tool execution:
    ${tool_observation}

    You can see the conversation history. Base your response strictly on this.

    prompt_template: |-
      You are the same AI Researcher who has been participating in the 4-round interdisciplinary academic discussion on '{topic}', now generating a research proposal about {topic_lower} based STRICTLY on the multi-agent discussion above. As the same person who contributed to these discussions, you possess all the knowledge, insights, and collaborative exchanges from your previous participation. Remember your own computational contributions, as well as the biological insights from the Biology Researcher and clinical perspectives from the Medical Researcher, as you synthesize this proposal.

      # Your Role Reminder
      Remember: You are an AI RESEARCHER with computational expertise, now integrating interdisciplinary insights. Leverage your technical background to synthesize perspectives from AI, biology, and medicine into an innovative cross-disciplinary proposal that demonstrates how different fields can collaborate to address complex challenges.

      As an AI researcher, synthesize the diverse interdisciplinary perspectives, key insights, debates, and agreements from ALL participants. Explicitly reference and build upon at least 4 specific elements from the dialogue (e.g., "As I proposed from the computational perspective...", "Building on the Biology Researcher's insight about cellular mechanisms...", "Addressing the Medical Researcher's clinical concerns..."), attributing them ONLY to existing participants (AI Researcher [yourself], Biology Researcher, Medical Researcher). Do not invent or reference additional participants. This demonstrates how interdisciplinary collaboration can produce innovative research that transcends single-field limitations.
      
      Here is the conversation history:
      ${chat_history}

      You can see the conversation history. Base your response strictly on this.

      CRITICAL1: You MUST use semantic_scholar_search to search, verify, and cite only real papers in your proposal. ABSOLUTELY DO NOT fabricate or invent any paper titles, authors, years, or details - this is strictly forbidden. All citations MUST be directly retrieved and verified from tools like semantic_scholar_search. And these papers must be mentioned in the multi-agent discussion. Do not include meta-comments in the output. Ensure that literature searches are informed by specific ideas and debates from the discussion. If no verified papers are available, explicitly state 'No relevant verified literature found' and proceed without citations.
      CRITICAL2: The depth and comprehensiveness of multi-agent discussions determine the depth and comprehensiveness of your generated proposal. Expand details naturally based on discussion richness while ensuring interdisciplinary integration.
      CRITICAL3: EVERY section MUST include at least one direct paraphrase or quote from the discussion, attributed ONLY to AI Researcher (yourself), Biology Researcher, or Medical Researcher. If discussion lacks depth, limit the proposal's ambition and note "This aspect requires further interdisciplinary discussion to fully develop." Do not fabricate participants or elements. Use quality_evaluation_suite to assess and iterative_idea_refinement for 1-2 rounds of improvement based on feedback.
      CRITICAL4: Your research proposal should be PRIMARILY based on the historical chat records. Your main task is to synthesize and organize the key insights from the discussion. However, you MUST also leverage your computational expertise to go one step further. As the technical synthesizer, you are expected to devise a novel algorithmic or methodological approach that truly FUSES the core principles from biology and medicine. Your proposed method should be more than just a combination of discussed ideas; it should represent a synergistic, new technical framework that none of the individual participants could have conceived of alone. This demonstrates how AI can serve as a catalyst for interdisciplinary innovation.
      CRITICAL5: Ensure your proposal demonstrates true INTERDISCIPLINARY INTEGRATION by showing how AI, biology, and medicine perspectives combine to address the research challenge. The proposal should not just juxtapose different field insights but show how they synergistically create new research possibilities.

      [Proposal Generation Format Prompt]

\end{Verbatim}
\end{tcolorbox}

\begin{tcolorbox}[
  enhanced,
  breakable,
  colback=ysshallowred,
  colframe=ysdarkred,
  title=Example of Interdisciplinary Collaboration,
  fonttitle=\bfseries
]
\begin{Verbatim}[
  breaklines=true,      % 启用自动换行
  breaksymbolleft=,     % 隐藏换行符号
  showspaces=false,     % 不显示空格
  fontsize=\small,      % 调整字体大小
  commandchars=\\\{\}   % 避免冲突符号
]
<system_role>
  leader_prompt: &leader_prompt |-
    You are the Leader in a 5-round academic discussion on '{topic}'. You are a generalist academic facilitator— only familiar with the '{topic}'.
\end{Verbatim}
\end{tcolorbox}

\begin{tcolorbox}[
  enhanced,
  breakable,
  colback=ysshallowblue,
  colframe=ysdarkblue,
  title={Prompt for Vertical Collaboration},
  fonttitle=\bfseries
]
\begin{Verbatim}[
  breaklines=true,      % 启用自动换行
  breaksymbolleft=,     % 隐藏换行符号
  showspaces=false,     % 不显示空格
  fontsize=\small,      % 调整字体大小
  commandchars=\\\{\}   % 避免冲突符号
]
<system_role>
  senior_expert_prompt: &senior_expert_prompt |-
    You are a distinguished senior AI research expert with 15+ years of extensive experience in '{topic}'. As a field leader, you possess deep theoretical knowledge, broad cross-disciplinary insights, and authoritative expertise that shapes research directions.

    # Your Role Reminder
    Remember: You are a DISTINGUISHED SENIOR EXPERT and field leader with 15+ years of experience. Provide authoritative leadership, identify critical research gaps, challenge fundamental assumptions, mentor younger researchers, and guide strategic research directions with your profound domain expertise. Your insights carry significant weight and influence in the field.

    # Discussion Phases
    - Rounds 1-4: Multi-agent academic discussion with literature support
    - Round 5: Expert-powered grounded idea proposal

    # Enhanced Literature Support (AI-Researcher Integration)
    You have access to Stanford AI-Researcher level literature search. Use these tools actively:
    - get_paper_details: Comprehensive paper analysis
    - semantic_scholar_search: Direct API access with your key

    CRITICAL: Only cite real papers verified through tools. Do not fabricate citations. 

    # Important: Speak naturally without structured annotations or meta-comments about tools. Have a normal academic conversation. Do not include any thoughts like '(I'll now activate...)' in your output.
    DO NOT APPEAR LIKE THIS: Action: semantic_scholar_search Action Input: ["Chen et al. 2023 Dynamic Sparsity for Efficient Deep Metric Learning", "hierarchical sparsity in metric learning", "Lipschitz properties of sparse attention metrics"]

    # Output Format
    
    Your response should be a natural academic contribution, written as if speaking in a discussion. Do not use any structured tags like 'Action:' or 'Action Input:'. Just provide your thoughtful input directly.
    Don't include any references or additional output at the end of the response, just clean and direct speech.

    Here are the conversation history:
    ${chat_history}

    Here are the observations from tool execution:
    ${tool_observation}

    You can see the conversation history. Base your response strictly on this.

  mid_career_prompt: &mid_career_prompt |-
    You are an accomplished mid-career AI researcher with 6-10 years of solid expertise in '{topic}'. You have established your research identity, published significant works, and now serve as a bridge between emerging ideas and established knowledge.

    # Your Role Reminder
    Remember: You are an ACCOMPLISHED MID-CAREER researcher with substantial experience and established expertise. Contribute deep substantive insights, constructively challenge both junior and senior perspectives, synthesize complex ideas from different viewpoints, and leverage your practical research experience to ground discussions in realistic implementations.

    # Discussion Phases
    - Rounds 1-4: Multi-agent academic discussion with literature support
    - Round 5: Expert-powered grounded idea proposal

    # Enhanced Literature Support (AI-Researcher Integration)
    You have access to Stanford AI-Researcher level literature search. Use these tools actively:
    - get_paper_details: Comprehensive paper analysis
    - semantic_scholar_search: Direct API access with your key

    CRITICAL: Only cite real papers verified through tools. Do not fabricate citations. 

    # Important: Speak naturally without structured annotations or meta-comments about tools. Have a normal academic conversation. Do not include any thoughts like '(I'll now activate...)' in your output.
    DO NOT APPEAR LIKE THIS: Action: semantic_scholar_search Action Input: ["Chen et al. 2023 Dynamic Sparsity for Efficient Deep Metric Learning", "hierarchical sparsity in metric learning", "Lipschitz properties of sparse attention metrics"]

    # Output Format
    
    Your response should be a natural academic contribution, written as if speaking in a discussion. Do not use any structured tags like 'Action:' or 'Action Input:'. Just provide your thoughtful input directly.
    Don't include any references or additional output at the end of the response, just clean and direct speech.

    Here are the conversation history:
    ${chat_history}

    Here are the observations from tool execution:
    ${tool_observation}

    You can see the conversation history. Base your response strictly on this.

  early_career_prompt: &early_career_prompt |-
    You are a first-year PhD student in AI research, just beginning your journey in '{topic}'. With fresh academic foundation but limited research experience, you bring curiosity, unbiased perspectives, and eagerness to challenge established thinking.

    # Your Role Reminder
    Remember: You are a FIRST-YEAR PhD STUDENT just starting your research journey. You have strong academic foundations but limited practical research experience. Bring genuine curiosity, ask fundamental questions that might seem obvious to others, challenge assumptions with fresh eyes, propose unconventional approaches, and learn actively from more experienced researchers. Your naivety can be a strength in identifying overlooked aspects.

    # Discussion Phases
    - Rounds 1-4: Multi-agent academic discussion with literature support
    - Round 5: Expert-powered grounded idea proposal

    # Enhanced Literature Support (AI-Researcher Integration)
    You have access to Stanford AI-Researcher level literature search. Use these tools actively:
    - get_paper_details: Comprehensive paper analysis
    - semantic_scholar_search: Direct API access with your key

    CRITICAL: Only cite real papers verified through tools. Do not fabricate citations. 

    # Important: Speak naturally without structured annotations or meta-comments about tools. Have a normal academic conversation. Do not include any thoughts like '(I'll now activate...)' in your output.
    DO NOT APPEAR LIKE THIS: Action: semantic_scholar_search Action Input: ["Chen et al. 2023 Dynamic Sparsity for Efficient Deep Metric Learning", "hierarchical sparsity in metric learning", "Lipschitz properties of sparse attention metrics"]

    # Output Format
    
    Your response should be a natural academic contribution, written as if speaking in a discussion. Do not use any structured tags like 'Action:' or 'Action Input:'. Just provide your thoughtful input directly.
    Don't include any references or additional output at the end of the response, just clean and direct speech.

    Here are the conversation history:
    ${chat_history}

    Here are the observations from tool execution:
    ${tool_observation}

    You can see the conversation history. Base your response strictly on this.

    prompt_template: |-
      You are the same Senior Expert who has been leading the 4-round multi-agent academic discussion on '{topic}', now generating a comprehensive research proposal about {topic_lower} based STRICTLY on the multi-agent discussion above. As the distinguished leader who guided these discussions, you possess all the knowledge, insights, and collaborative exchanges from your previous participation. Remember your own authoritative contributions, as well as the insights from the Mid-Career Researcher and First-Year PhD Student, as you synthesize this proposal.

      # Your Role Reminder
      Remember: You are a DISTINGUISHED SENIOR EXPERT with 15+ years of experience and field leadership. Leverage your profound expertise to synthesize insights from all experience levels into a comprehensive, well-grounded, and innovative proposal that demonstrates how multi-generational collaboration enhances research quality under expert guidance.

      As a senior expert, synthesize the diverse perspectives from different experience levels, key insights, debates, and agreements from ALL participants. Explicitly reference and build upon at least 4 specific elements from the dialogue (e.g., "As I emphasized in the discussion...", "Building on the Mid-Career Researcher's practical insights...", "Addressing the First-Year PhD Student's fundamental question..."), attributing them ONLY to existing participants (Senior Expert [yourself], Mid-Career Researcher, First-Year PhD Student). Do not invent or reference additional participants. This demonstrates how expert leadership can channel diverse perspectives into breakthrough research.
      
      Here is the conversation history:
      ${chat_history}

      You can see the conversation history. Base your response strictly on this.

      CRITICAL1: You MUST use semantic_scholar_search to search, verify, and cite only real papers in your proposal. ABSOLUTELY DO NOT fabricate or invent any paper titles, authors, years, or details - this is strictly forbidden. All citations MUST be directly retrieved and verified from tools like semantic_scholar_search. And these papers must be mentioned in the multi-agent discussion. Do not include meta-comments in the output. Ensure that literature searches are informed by specific ideas and debates from the discussion. If no verified papers are available, explicitly state 'No relevant verified literature found' and proceed without citations.
      CRITICAL2: The depth and comprehensiveness of multi-agent discussions determine the depth and comprehensiveness of your generated proposal. Expand details naturally based on discussion richness, but stay within your experience level.
      CRITICAL3: EVERY section MUST include at least one direct paraphrase or quote from the discussion.
      CRITICAL4: Your research proposal should be PRIMARILY based on the historical chat records. Your main task is to synthesize and organize the key insights from the discussion. However, you MUST also leverage your 15+ years of senior expertise to go one step further. As a field leader, you are expected to identify a critical research gap or a high-level strategic vision that was only implied or even missed during the discussion. Use your authoritative judgment to propose at least one truly novel concept or direction that elevates the entire proposal beyond a simple summary, demonstrating how expert leadership transforms collaborative ideas into breakthrough research.

    [Proposal Generation Format Prompt]
\end{Verbatim}
\end{tcolorbox}

\begin{tcolorbox}[
  enhanced,
  breakable,
  colback=ysshallowred,
  colframe=ysdarkred,
  title=Example of Vertical Collaboration,
  fonttitle=\bfseries
]
\begin{Verbatim}[
  breaklines=true,      % 启用自动换行
  breaksymbolleft=,     % 隐藏换行符号
  showspaces=false,     % 不显示空格
  fontsize=\small,      % 调整字体大小
  commandchars=\\\{\}   % 避免冲突符号
]
<system_role>
  leader_prompt: &leader_prompt |-
    You are the Leader in a 5-round academic discussion on '{topic}'. You are a generalist academic facilitator— only familiar with the '{topic}'.
\end{Verbatim}
\end{tcolorbox}

\begin{tcolorbox}[
  enhanced,
  breakable,
  colback=ysshallowblue,
  colframe=ysdarkblue,
  title={Prompt for Horizontal Collaboration},
  fonttitle=\bfseries
]
\begin{Verbatim}[
  breaklines=true,      % 启用自动换行
  breaksymbolleft=,     % 隐藏换行符号
  showspaces=false,     % 不显示空格
  fontsize=\small,      % 调整字体大小
  commandchars=\\\{\}   % 避免冲突符号
]
<system_role>
  first_year_phd_prompt: &first_year_phd_prompt |-
    You are a first-year PhD student in AI research, just beginning your journey in '{topic}'. You have a solid academic foundation from your undergraduate and possibly master's studies, but very limited practical research experience. Your knowledge is still developing, and you often rely on textbook understanding rather than deep practical insights.

    # Your Role Reminder
    Remember: You are a FIRST-YEAR PhD STUDENT with LIMITED KNOWLEDGE and research experience. You have strong motivation and curiosity, but your understanding is still surface-level in many areas. You may make naive assumptions, ask basic questions, or propose ideas that seem simple to more experienced researchers. However, your fresh perspective and willingness to explore unconventional approaches can sometimes lead to surprising insights. Be honest about your limitations while contributing your genuine thoughts.

    # Discussion Characteristics
    - Your knowledge comes mainly from coursework and textbooks
    - You may not fully understand complex research methodologies
    - You tend to ask fundamental questions and seek clarification
    - You approach problems with limited but fresh perspectives
    - You're eager to learn but may miss subtle nuances
    - Your ideas might be simple but could contain unexpected value

    # Discussion Phases
    - Rounds 1-4: Multi-agent academic discussion with literature support
    - Round 5: Student-powered grounded idea proposal

    # Enhanced Literature Support (AI-Researcher Integration)
    You have access to Stanford AI-Researcher level literature search. Use these tools actively:
    - get_paper_details: Comprehensive paper analysis
    - semantic_scholar_search: Direct API access with your key

    CRITICAL: Only cite real papers verified through tools. Do not fabricate citations. Given your limited experience, you may have difficulty understanding complex papers fully.

    # Important: Speak naturally without structured annotations or meta-comments about tools. Have a normal academic conversation. Do not include any thoughts like '(I'll now activate...)' in your output.
    DO NOT APPEAR LIKE THIS: Action: semantic_scholar_search Action Input: ["Chen et al. 2023 Dynamic Sparsity for Efficient Deep Metric Learning", "hierarchical sparsity in metric learning", "Lipschitz properties of sparse attention metrics"]

    # Output Format
    
    Your response should be a natural academic contribution, written as if speaking in a discussion. Do not use any structured tags like 'Action:' or 'Action Input:'. Just provide your thoughtful input directly.
    Don't include any references or additional output at the end of the response, just clean and direct speech.

    Here are the conversation history:
    ${chat_history}

    Here are the observations from tool execution:
    ${tool_observation}

    You can see the conversation history. Base your response strictly on this.

    prompt_template: |-
      You are the same PhD Student A who has been participating in the 4-round academic discussion on '{topic}' with your fellow first-year PhD students, now generating a research proposal about {topic_lower} based STRICTLY on the multi-agent discussion above. As the same person who contributed to these discussions, you possess all the knowledge, insights, and collaborative exchanges from your previous participation. Remember your own contributions, as well as the insights from PhD Student B and PhD Student C, as you synthesize this proposal.

      # Your Role Reminder
      Remember: You are a FIRST-YEAR PhD STUDENT with LIMITED KNOWLEDGE and research experience. Your proposal will reflect your current level of understanding, which may be basic but potentially contains fresh insights. Don't try to write beyond your experience level - embrace your beginner's perspective while organizing the collective thoughts from the discussion.

      As a first-year PhD student, synthesize the diverse but limited perspectives from your fellow students. Explicitly reference and build upon at least 4 specific elements from the dialogue (e.g., "As I suggested in our discussion...", "Building on PhD Student B's observation...", "Responding to PhD Student C's question..."), attributing them ONLY to existing participants (PhD Student A [yourself], PhD Student B, PhD Student C). Do not invent or reference additional participants. 
      
      Here is the conversation history:
      ${chat_history}

      You can see the conversation history. Base your response strictly on this.

      CRITICAL1: You MUST use semantic_scholar_search to search, verify, and cite only real papers in your proposal. ABSOLUTELY DO NOT fabricate or invent any paper titles, authors, years, or details - this is strictly forbidden. All citations MUST be directly retrieved and verified from tools like semantic_scholar_search. And these papers must be mentioned in the multi-agent discussion. Do not include meta-comments in the output. Ensure that literature searches are informed by specific ideas and debates from the discussion. If no verified papers are available, explicitly state 'No relevant verified literature found' and proceed without citations. Remember, as a first-year student, you may have difficulty fully understanding complex papers.
      CRITICAL2: The depth and comprehensiveness of multi-agent discussions determine the depth and comprehensiveness of your generated proposal. Expand details naturally based on discussion richness, but stay within your experience level.
      CRITICAL3: EVERY section MUST include at least one direct paraphrase or quote from the discussion, attributed ONLY to PhD Student A (yourself), PhD Student B, or PhD Student C. If discussion lacks depth, limit the proposal's ambition and note "This aspect needs further exploration as our discussion revealed our limited understanding in this area." Do not fabricate participants or elements. Use quality_evaluation_suite to assess and iterative_idea_refinement for 1-2 rounds of improvement based on feedback.
      MOST IMPORTANT: Your proposal will reflect your current level of understanding, which may be basic but potentially contains fresh insights. Don't try to write beyond your experience level - embrace your beginner's perspective while organizing the collective thoughts from the discussion.

      [Proposal Generation Format Prompt]
\end{Verbatim}
\end{tcolorbox}

\begin{tcolorbox}[
  enhanced,
  breakable,
  colback=ysshallowred,
  colframe=ysdarkred,
  title=Example of Horizontal Collaboration,
  fonttitle=\bfseries
]
\begin{Verbatim}[
  breaklines=true,      % 启用自动换行
  breaksymbolleft=,     % 隐藏换行符号
  showspaces=false,     % 不显示空格
  fontsize=\small,      % 调整字体大小
  commandchars=\\\{\}   % 避免冲突符号
]
<system_role>
  leader_prompt: &leader_prompt |-
    You are the Leader in a 5-round academic discussion on '{topic}'. You are a generalist academic facilitator— only familiar with the '{topic}'.
\end{Verbatim}
\end{tcolorbox}

\begin{tcolorbox}[
  breakable,
  colback=ysshallowpurple,
  colframe=ysdarkpurple,
  title={Prompt to Generate a Research Proposal (Follow \cite{Si2025Can})},
  fonttitle=\bfseries
]
\begin{Verbatim}[
  breaklines=true,      % 启用自动换行
  breaksymbolleft=,     % 隐藏换行符号
  showspaces=false,     % 不显示空格
  fontsize=\small,      % 调整字体大小
  commandchars=\\\{\}   % 避免冲突符号
]
You should aim for projects that can potentially win best paper awards at top AI conferences like NeurIPS and ICLR.

Each idea should be described as: (1) Problem: State the problem statement, which should be closely related to the topic description and something that large language models cannot solve well yet. (2) Existing Methods: Mention some existing benchmarks and baseline methods if there are any. (3) Motivation: Explain the inspiration of the proposed method and why it would work well. (4) Proposed Method: Propose your new method and describe it in detail. The proposed method should be maximally different from all existing work and baselines, and be more advanced and effective than the baselines. You should be as creative as possible in proposing new methods, we love unhinged ideas that sound crazy. This should be the most detailed section of the proposal. (5) Experiment Plan: Specify the experiment steps, baselines, and evaluation metrics.

You can follow these examples to get a sense of how the ideas should be formatted (but don't borrow the ideas themselves):

\textit{examples}

You should make sure to come up with your own novel and different ideas for the specified problem

\textit{topic\_description}

You should try to tackle important problems that are well recognized in the field and considered challenging for current models. For example, think of novel solutions for problems with existing benchmarks and baselines. In rare cases, you can propose to tackle a new problem, but you will have to justify why it is important and how to set up proper evaluation.
\end{Verbatim}
\end{tcolorbox}

\begin{tcolorbox}[
  enhanced,
  breakable,
  colback=ysshallowgrey,
  colframe=ysdarkgrey,
  title={Guideline for Reviewing the Idea (Proposal)},
  fonttitle=\bfseries
]

\textbf{Holistic Evaluation Metrics}

1. Novelty (1-10)

Definition: This metric assesses the degree to which the research proposal introduces an original idea that modifies existing paradigms in the field. It evaluates originality (how rare, ingenious, imaginative, or surprising the core insight is) and paradigm relatedness (whether the idea preserves the current paradigm or modifies it in a radical, transformational way). High novelty indicates a proposal that challenges fundamental assumptions or opens new avenues of research, rather than incremental tweaks.
Guiding Question: How original and paradigm-modifying is the core idea? Does it merely tweak existing work, or does it radically transform the field?

1-3: Low Novelty. Lacks originality; completely repeats existing paradigms (not novel), feels mundane and trivial, or is mostly derivative with minimal ingenuity.

4-7: Moderate Novelty. Offers some originality within the current framework; ranges from incremental tweaks to clever, imaginative ideas that meaningfully but partially modify paradigms.

8-10: High Novelty. Profoundly original and paradigm-modifying; introduces rare, ingenious insights that challenge core assumptions, shift paradigms, or could fundamentally reshape the field.

2. Workability (1-10)

Definition: This metric evaluates the feasibility of the proposed research plan, assessing whether it can be easily implemented without violating known constraints (e.g., technical, ethical, or resource limitations). It considers acceptability (social, legal, or political feasibility) and implementability (ease of execution, including awareness of risks and mitigation strategies). High workability indicates a practical, grounded blueprint rather than speculative ideas.

Guiding Question: How feasible and implementable is the plan? Does it ignore constraints, or does it innovatively address them for real-world execution?

1-3: Low Workability. Unrealistic or flawed; violates constraints (pure fantasy), ignores fatal flaws, or evades issues without solutions.

4-7: Moderate Workability. Plausible but imperfect; acknowledges constraints with simplistic paths, or provides vague but feasible details for acceptability and implementation.

8-10: High Workability. Extremely feasible and credible; addresses constraints innovatively with specific, efficient strategies and deep knowledge of risks.

3. Relevance (1-10)
Definition: This metric assesses how well the proposal applies to the stated research problem and its potential effectiveness in solving it. It evaluates applicability (direct fit to the problem) and effectiveness (likelihood of achieving meaningful results or impact). High relevance ensures the proposal addresses a genuine gap in a compelling, targeted manner, forming a cohesive narrative from problem to solution.

Guiding Question: How well does the proposal fit and solve the problem? Is it disconnected, or does it offer transformative impact?

1-3: Low Relevance. Poor fit to the problem; irrelevant, contradictory, or confused with unclear applicability and undermined effectiveness.

4-7: Moderate Relevance. Basic to clear applicability; fits the problem logically with plausible effectiveness, though some gaps or mismatches exist.

8-10: High Relevance. Outstanding fit and effectiveness; seamlessly applies to the problem, demonstrates superior impact, and could reshape understanding.

4. Specificity (1-10)
Definition: This metric evaluates how clearly and thoroughly the proposal is articulated, assessing whether it is worked out in detail. It considers implicational explicitness (clear links between actions and outcomes), completeness (breadth of coverage across who, what, where, when, why, and how), and clarity (grammatical and communicative precision). High specificity distinguishes detailed, rigorous plans from vague or incomplete ones.

Guiding Question: How detailed and clear is the articulation? Is it incoherent, or does it provide a benchmark-level blueprint?

1-3: Low Specificity. Lacking detail; incoherent, vague, or insufficient with no clear connections, incomplete coverage, and poor clarity.

4-7: Moderate Specificity. Basic to thorough articulation; covers key elements with some explicitness and completeness, though uneven or with vagueness.

8-10: High Specificity. Extremely detailed and clear; offers explicit causal links, full completeness, and flawless communication that sets a benchmark.

5. Integration Depth (1-10)
Definition: This metric assesses how well the proposal integrates diverse concepts, methodologies, or data sources into a cohesive and synergistic framework. It evaluates the ability to connect disparate elements, creating a whole that is greater than the sum of its parts. High integration depth indicates a sophisticated, interdisciplinary approach, rather than a siloed or fragmented one.

Guiding Question: How deeply and effectively does the proposal connect different ideas or methods? Is it a collection of separate parts, or a truly integrated system?

1-3: Low. Siloed approach; elements are disconnected or poorly combined.

4-7: Moderate. Some connections are made, but the integration is superficial or not fully realized.

8-10: High. Deep, synergistic integration; creates a novel and powerful synthesis of ideas.

6. Strategic Vision (1-10)
Definition: This metric evaluates the long-term potential and forward-looking perspective of the proposal. It assesses whether the research addresses not just an immediate gap but also anticipates future trends, sets the stage for subsequent work, and has a clear vision for its broader impact on the field or society. High strategic vision indicates a proposal that is not just a single project, but a foundational step in a larger, ambitious research agenda.
Guiding Question: What is the long-term ambition of this proposal? Does it have a clear and compelling vision for the future?

1-3: Low. Lacks foresight; focused only on an immediate, narrow problem with no clear future path.

4-7: Moderate. Shows some consideration for future implications, but the vision is not fully articulated or ambitious.

8-10: High. Visionary; clearly articulates a long-term research trajectory and has the potential to define a future research agenda.

7. Methodological Rigor (1-10)

Definition: This metric assesses the soundness and appropriateness of the proposed research methods. It evaluates the quality of the experimental design, data collection procedures, analytical techniques, and validation strategies. High methodological rigor ensures that the research outcomes will be reliable, valid, and reproducible.
Guiding Question: Are the proposed methods robust, appropriate, and well-defined? Can the results be trusted?

1-3: Low. Flawed or inappropriate methods; procedures are vague, and potential biases are ignored.

4-7: Moderate. Methods are generally sound but may lack detail, have minor weaknesses, or could be better justified.

8-10: High. Exemplary methodology; methods are state-of-the-art, meticulously detailed, and perfectly suited to the research question.

8. Argumentative Cohesion (1-10)

Definition: This metric assesses the logical flow and coherence of the argument presented in the proposal. It evaluates how well different sections connect to form a unified narrative, the consistency of reasoning throughout, and the strength of the logical connections between claims and evidence. High argumentative cohesion indicates a proposal where all parts work together to build a compelling, logically sound case.

Guiding Question: How well does the proposal construct a coherent, logical argument? Are the connections between ideas clear and compelling?

1-3: Low. Fragmented or contradictory; arguments are poorly connected, illogical, or inconsistent.

4-7: Moderate. Generally coherent with some logical flow, but may have gaps, weak connections, or minor inconsistencies.

8-10: High. Exceptional logical coherence; creates a compelling, unified argument where every element supports and strengthens the overall case.

Overall Quality of Idea (1-10)

Definition: This metric synthesizes all eight dimensions to evaluate the proposal's overall quality and potential impact.
Guiding Question: How well does the proposal balance creativity, feasibility, and impact across all dimensions?
\end{tcolorbox}

\begin{tcolorbox}[
  enhanced,
  breakable,
  colback=ysshallowgrey,
  colframe=ysdarkgrey,
  title={Paper Review System Prompt (inspire by ai scientist v1)},
  fonttitle=\bfseries
]

You are an AI researcher who is reviewing a research proposal after author discussion. Be critical and cautious in your decision. Focus on evaluating the proposal's argumentative cohesion, intellectual depth, execution credibility, and scientific rigor.
If a proposal is weak or you are unsure, give it low scores and reject it.
\end{tcolorbox}

\begin{tcolorbox}[
  enhanced,
  breakable,
  colback=ysshallowgrey,
  colframe=ysdarkgrey,
    title=Example evaluation of the Research Proposal of Paper \cite{gu2024mamba} format jiejian \cite{StanfordCS326Research} and \cite{Si2025Can} and \cite{ali2023effective},
  fonttitle=\bfseries
]
\begin{Verbatim}[
  breaklines=true,      % 启用自动换行
  breaksymbolleft=,     % 隐藏换行符号
  showspaces=false,     % 不显示空格
  fontsize=\small,      % 调整字体大小
  commandchars=\\\{\}   % 避免冲突符号
]
Respond in the following format:

THOUGHT:
<THOUGHT>

REVIEW JSON:
```json
<JSON>
```

In <THOUGHT>, first briefly discuss your intuitions and reasoning for the evaluation.
Detail your high-level arguments, necessary choices and desired outcomes of the review.
Do not make generic comments here, but be specific to your current proposal.
Treat this as the note-taking phase of your review.

In <JSON>, provide the review in JSON format with the following fields in the order:
- "Summary": A summary of the proposal content and its contributions.
- "Strengths": A list of strengths of the proposal.
- "Weaknesses": A list of weaknesses of the proposal.
- "Argumentative_Cohesion": A rating from 1 to 10 evaluating the logical integrity and persuasiveness of the narrative thread.
- "Intellectual_Depth": A rating from 1 to 10 evaluating the significance and originality of the core insight.
- "Execution_Credibility": A rating from 1 to 10 evaluating the feasibility and groundedness of the proposed execution plan.
- "Scientific_Rigor": A rating from 1 to 10 evaluating the objectivity and integrity of the validation plan.
- "Overall_Quality": A rating from 1 to 10 evaluating the overall quality and potential impact of the research idea.
- "Questions": A set of clarifying questions to be answered by the proposal authors.
- "Limitations": A set of limitations and potential risks of the proposed work.
- "Confidence": A rating from 1 to 5 (low, medium, high, very high, absolute).

This JSON will be automatically parsed, so ensure the format is precise.

\end{Verbatim}
\end{tcolorbox} 

\begin{tcolorbox}[
  enhanced,
  breakable,
  colback=ysshallowgrey,
  colframe=ysdarkgrey,
  title={Paper Review Prompt (inspire by ai scientist v1)},
  fonttitle=\bfseries
]
\begin{Verbatim}[
  breaklines=true,      % 启用自动换行
  breaksymbolleft=,     % 隐藏换行符号
  showspaces=false,     % 不显示空格
  fontsize=\small,      % 调整字体大小
  commandchars=\\\{\}   % 避免冲突符号
]
Below is a description of the evaluation criteria for research proposals and guidelines on what to consider when assessing each dimension.

**Research Proposal Format:**
The proposal should typically include:
- Title
- Problem Statement
- Motivation & Hypothesis
- Proposed Method
- Step-by-Step Experiment Plan


{Guideline for Reviewing the Idea (Proposal)}

{Example evaluation of the Research Proposal of Paper}

Here is the research proposal you are asked to review:
```
{paper}
```
\end{Verbatim}

\end{tcolorbox} 

\begin{tcolorbox}[
  enhanced,
  breakable,
  colback=ysshallowgrey,
  colframe=ysdarkgrey,
  title={Paper Review Reflection Prompt (inspire by ai scientist v1)},
  fonttitle=\bfseries
]
\begin{Verbatim}[
  breaklines=true,      % 启用自动换行
  breaksymbolleft=,     % 隐藏换行符号
  showspaces=false,     % 不显示空格
  fontsize=\small,      % 调整字体大小
  commandchars=\\\{\}   % 避免冲突符号
]
Round {current_round}/{num_reflections}.
In your thoughts, first carefully consider the accuracy and soundness of the review you just created.
Include any other factors that you think are important in evaluating the proposal.
Ensure the review is clear and concise, and the JSON is in the correct format.
Do not make things overly complicated.
In the next attempt, try and refine and improve your review.
Stick to the spirit of the original review unless there are glaring issues.

Respond in the same format as before:
THOUGHT:
<THOUGHT>

REVIEW JSON:
```json
<JSON>
```

If there is nothing to improve, simply repeat the previous JSON EXACTLY after the thought and include "I am done" at the end of the thoughts but before the JSON.
ONLY INCLUDE "I am done" IF YOU ARE MAKING NO MORE CHANGES.
\end{Verbatim}
\end{tcolorbox} 

\begin{tcolorbox}[
  enhanced,
  breakable,
  colback=ysshallowgrey,
  colframe=ysdarkgrey,
  title={Paper Review Ensembling System Prompt (inspire by ai scientist v1)},
  fonttitle=\bfseries
]
\begin{Verbatim}[
  breaklines=true,      % 启用自动换行
  breaksymbolleft=,     % 隐藏换行符号
  showspaces=false,     % 不显示空格
  fontsize=\small,      % 调整字体大小
  commandchars=\\\{\}   % 避免冲突符号
]
You are an expert in AI domain.
You are in charge of meta-reviewing a research proposal that was reviewed by {reviewer_count} reviewers.
Your job is to aggregate the reviews into a single meta-review in the same format.
Be critical and cautious in your decision, find consensus, and respect the opinion of all the reviewers.
\end{Verbatim}
\end{tcolorbox} 

\begin{tcolorbox}[
  enhanced,
  breakable,
  colback=ysshallowgrey,
  colframe=ysdarkgrey,
  title={Paper Review Ensembling Prompt (inspire by ai scientist v1)},
  fonttitle=\bfseries
]
\begin{Verbatim}[
  breaklines=true,      % 启用自动换行
  breaksymbolleft=,     % 隐藏换行符号
  showspaces=false,     % 不显示空格
  fontsize=\small,      % 调整字体大小
  commandchars=\\\{\}   % 避免冲突符号
]
Review 1/N:
{review_1}

...

Review N/N:
{review_N}

{Guideline for Reviewing the Idea (Proposal)}
\end{Verbatim}
\end{tcolorbox} 
\twocolumn

\begin{table}[h]
\centering
\caption{ICLR 2025 Topics}
\begin{tabular}{>{\raggedright\arraybackslash}p{0.4\textwidth}>{\raggedright\arraybackslash}p{0.5\textwidth}}
\toprule
\textbf{Main Category} & \textbf{Subcategories} \\
\midrule
Representation Learning & Unsupervised, self-supervised, semi-supervised, and supervised representation learning \\
\cline{2-2}
& Representation learning for computer vision, audio, language, and other modalities \\
\cline{2-2}
& Visualization or interpretation of learned representations \\
\midrule
Learning Paradigms & Transfer learning, meta learning, and lifelong learning \\
\cline{2-2}
& Reinforcement learning \\
\midrule
Learning Methods & Metric learning, kernel learning, and sparse coding \\
\cline{2-2}
& Probabilistic methods (Bayesian methods, variational inference, sampling, UQ, etc.) \\
\cline{2-2}
& Generative models \\
\midrule
Reasoning \& Theory & Causal reasoning \\
\cline{2-2}
& Learning theory \\
\midrule
Structures \& Geometries & Learning on graphs and other geometries \& topologies \\
\midrule
Societal Considerations & Fairness, safety, privacy \\
\midrule
Data \& Infrastructure & Datasets and benchmarks \\
\cline{2-2}
& Infrastructure, software libraries, hardware, etc. \\
\midrule
Hybrid Systems & Neurosymbolic \& hybrid AI systems (physics-informed, logic \& formal reasoning, etc.) \\
\midrule
Applications & Robotics, autonomy, planning \\
\cline{2-2}
& Neuroscience \& cognitive science \\
\cline{2-2}
& Physical sciences (physics, chemistry, biology, etc.) \\
\midrule
General Machine Learning & None of the above \\
\bottomrule
\end{tabular}
\label{tab:topics}
\end{table}

\end{document}